\pdfoutput=1

\documentclass{article}

\usepackage[final]{neurips_2022}

\usepackage[utf8]{inputenc} 
\usepackage[T1]{fontenc}    
\usepackage[hidelinks]{hyperref}  
\usepackage{xcolor}
\hypersetup{
  colorlinks,
  citecolor=blue,
  linkcolor=blue,
  urlcolor=blue}

\usepackage{url}            
\usepackage{booktabs}       
\usepackage{amsfonts}       
\usepackage{nicefrac}       
\usepackage{microtype}       
\usepackage[tikz]{bclogo}
\usepackage{lipsum}
\usepackage{times}
\usepackage{latexsym}
\usepackage{algpseudocode}

\usepackage[T1]{fontenc}

\usepackage[utf8]{inputenc}

\usepackage{microtype}

\usepackage{booktabs}
\usepackage{graphicx} 
\usepackage{graphics}
\usepackage{amssymb}
\usepackage{amsmath}
\usepackage{pifont}
\usepackage{makecell}
\usepackage{wrapfig}
\usepackage{resizegather}
\usepackage{etoolbox}
\usepackage{booktabs,makecell,tabularx} 
\usepackage{enumitem}
\usepackage{mathtools}
\usepackage{cleveref}
\usepackage{multirow}
\usepackage{booktabs,array}
\usepackage{amsmath, bm}
\usepackage{color, colortbl}
\usepackage{xcolor}
\usepackage{colortbl}
\usepackage{mathtools}
\DeclarePairedDelimiter\ceil{\lceil}{\rceil}

\usepackage[ruled,vlined]{algorithm2e}         
\newcolumntype{P}[1]{>{\centering\arraybackslash}p{#1}}

\let\oldnl\nl
\definecolor{Gray}{gray}{0.9}
\definecolor{LightCyan}{rgb}{0.88,1,1}
\newcommand{\nonl}{\renewcommand{\nl}{\let\nl\oldnl}}
\newcommand{\comm}[1]{{\nonl{\small{{\color{brown}{/*~#1}~*/}}}}}

\newcolumntype{H}{>{\setbox0=\hbox\bgroup}c<{\egroup}@{}}

\newcommand{\relook}[1]{\textcolor{blue}{}}

\definecolor{Gray}{gray}{0.85}
\definecolor{LightCyan}{rgb}{0.88,1,1}
 
\newcolumntype{a}{>{\columncolor{Gray}}c}
\newcolumntype{b}{>{\columncolor{white}}c}

\title{Unsupervised Cross-Task Generalization \\ via Retrieval Augmentation}

\author{%
  Bill Yuchen Lin$^\dag$\quad
Kangmin Tan$^{\dag}$~\quad Chris Miller$^{\dag}$\quad Beiwen Tian$^\ddag$\quad Xiang Ren$^\dag$ \\
$^\dag$ University of Southern California \qquad 
$^\ddag$ Tsinghua University\\
\texttt{\{yuchen.lin,kangmint,millercs,xiangren\}@usc.edu}\\ 
}

\begin{document}

\maketitle

\begin{abstract}
Humans can perform unseen tasks by recalling relevant skills acquired previously and then generalizing them to the target tasks, even if there is no supervision at all.
In this paper, we aim to improve this kind of cross-task generalization ability of massive multi-task language models, such as T0 and FLAN, in an unsupervised setting.
We propose a retrieval-augmentation method named ReCross that takes a few \textit{unlabelled} examples as queries to retrieve a small subset of upstream data and uses them to update the multi-task model for better generalization.
ReCross is a straightforward yet effective retrieval method that combines both efficient dense retrieval and effective pair-wise reranking.
Our results and analysis show that it significantly outperforms both non-retrieval methods and other baseline methods.
\footnote{Our data, code, and supplementary materials are at   
\url{https://inklab.usc.edu/ReCross/}.
}

\end{abstract}
\section{Introduction}
\label{sec:intro} 

Advances in pre-training techniques for large language models (LMs) have considerably improved natural language processing (NLP) models on various important tasks via fine-tuning with labeled data.
While these fine-tuned models are impressive in their target tasks, they can hardly generalize to unseen tasks. 
This thus makes it difficult to approach the general linguistic intelligence that we ultimately want an NLP model to enjoy.
A promising avenue is to train a massively multi-task model that learns a large set of NLP tasks.
However, in real-world applications, users often expect a multi-task NLP model can also perform unseen tasks that they are interested in.
These users may only be able to provide a few \textit{unlabeled} examples (i.e., the input-only data) of the target tasks with natural-language instructions. 
How can we generalize the multi-task model to unseen tasks without labels?
This desirable ability is dubbed ``unsupervised cross-task generalization.''

Recent studies show that multi-task prompted training makes language models better in cross-task generalization, especially when  natural-language instructions are used for formatting the training data~\citep{ye-etal-2021-crossfit, sanh2021t0, Wei2021FinetunedLM}.
The general recipe is to first fine-tune a text-to-text language model such as T5~\citep{t5} on a multi-task mixture of diverse NLP datasets that are converted to sequence-to-sequence formats.
We use the term \textit{upstream learning}  to refer to this multi-task training stage.
Given a target task that is unseen during upstream learning, we want the upstream multi-task model to also perform well on it via reusing the  previously acquired knowledge.
FLAN~\citep{Wei2021FinetunedLM} and T0~\citep{sanh2021t0} both use natural language (NL) instructions as prompts to reformat the data of various NLP tasks for upstream learning and generalization.
Their results suggest that NL instructions are keys to unsupervised cross-task generalization.

Despite of the exciting results from~\cite{Wei2021FinetunedLM} and \cite{sanh2021t0}, their studies are limited to the analysis of the task generalization performance of the \textit{frozen, target-agnostic} upstream models (i.e., FLAN and T0).
We argue that the generalization performance can be further improved if we can exploit the unlabeled data of target tasks as hints for adjusting the  upstream model to a more dedicated, target-aware model.
Intuitively, the upstream examples that share similar skills with the target  task should help the task generalization if the upstream model could recap these skills via retrieving.
Motivated by this idea, 
we propose to further improve the cross-task generalization ability of upstream models via \textit{retrieval augmentation} from the upstream data. 

The key challenge of such retrieval augmentation is to predict the \textit{example-level utility} for cross-task generalization, which we introduce with details in Sec.~\ref{sec:problem}.
To address the challenges, we present a two-stage retrieval-augmentation framework, ReCross, for unsupervised cross-task generalization in Section~\ref{sec:methods}.
Specifically, we pre-compute a dense index by encoding all upstream data as dense vectors.
Given a set of unlabeled examples, we first use them to retrieve an initial set of upstream data by using encoded queries to efficiently search over the dense index.
Then, we apply the reranking module to carefully analyze the utility of each candidate example.
To get such a reranker, we fine-tune a cross-encoder model with distant supervision mined by a novel algorithm.
Finally, we take top-ranking retrieved data to fine-tune the upstream model for a few steps and use this updated model for inference on the target task in the future (i.e., the retrieval augmentation and model update is a one-time procedure for each unseen task).

To more efficiently evaluate generalization methods without losing the generality, we train a variant of T0-like models, named BART0, which has comparable performance with T0-3B yet is 8x smaller. 
Our extensive experiments show that the proposed ReCross outperforms the baseline methods by a large margin. For example, ReCross improves the non-retrieval methods by 4 points on the overall performance of 10 target tasks and similarly on a few BigBench tasks. 
We also analyze the distribution of the retrieved data to understand the behavior of retrieval-augmentation methods better and find that ReCross has a very different distribution compared to semantic retrieval baselines.

\begin{figure*}[t]
	\centering
    \vspace{-1em}
	\includegraphics[width=0.9\linewidth]{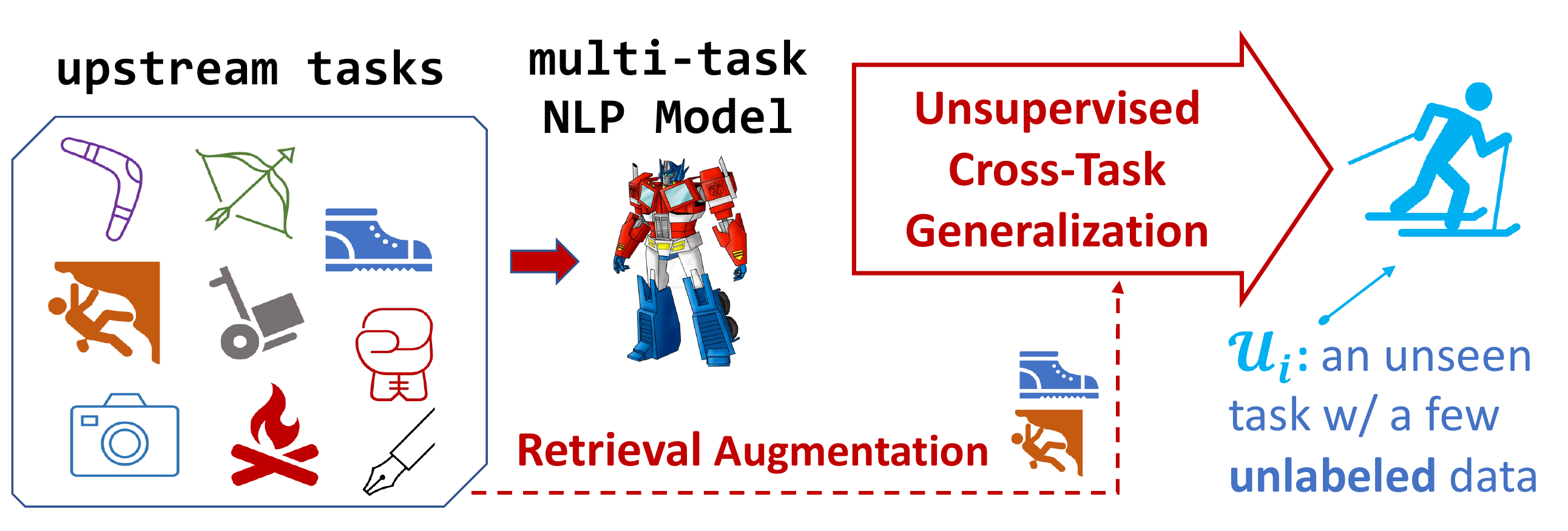} 
	\caption{\textbf{The unsupervised cross-task generalization problem.} In the upstream training stage, we train a  multi-task NLP model, $\mathcal{M}$, with a diverse collection of upstream tasks. In the generalization stage, given an unseen task $\mathcal{U}_i$ with a few unlabeled examples $Q_i$, we want to update the upstream model (via retrieval augmentation) such that it can generalize to the target task. \relook{(\textbf{updated!})} }
	\label{fig:intro}
\end{figure*}

\section{Problem Formulation}
\label{sec:problem}

\paragraph{Massively Multi-Task Language Models.}
To build a general NLP model that can serve a wide range of real-world downstream applications,
it is important to train a massively multi-task upstream model.
We assume there are $N$ different \textbf{upstream tasks} (e.g., sentiment analysis of IMDB reviews), dubbed as $\{\mathcal{T}_1, \dots, \mathcal{T}_N\}$.
We use ${D}$ to denote the collection of all labeled data  for these upstream tasks (i.e., the \textbf{upstream data}), which are then used for training a massive multi-task model $\mathcal{M}$ (e.g., BART, T5, and other Transformer-based models).
The datasets of these upstream tasks are all converted to a shared \textit{text-to-text} format using natural-language instruction templates such as PromptSource~\citep{bach2022promptsource} to reformat data of different NLP tasks. 
This pipeline has become a common approach, adopted by several recent massive multi-task models for NLP, such as T0~\citep{sanh2021t0}, FLAN~\citep{Wei2021FinetunedLM}, and CrossFit~\citep{ye-etal-2021-crossfit}. 

\paragraph{Unsupervised Cross-Task Generalization. }
In real-world scenarios, it is very common that users to want a general multi-task model to perform tasks of their interest, even if their target tasks are never seen before by the upstream model.
For these unseen target tasks, users usually can provide only a few \textit{unlabeled} examples (i.e., the input-only data) of them for specifying the task instructions. 
This is the reason why we need to study how to generalize a multi-task LM to unseen tasks with only a few \textit{unlabeled} examples, i.e., \textit{unsupervised cross-task generalization}.
For instance, in Fig.~\ref{fig:intro}, the \textit{unseen task} $\mathcal{U}_i$ is a coreference-resolution task that is not covered by the upstream training (the top-right box in Fig.~\ref{fig:intro}).
We have only a few inputs for it as the ``\textit{hints}'' for cross-task generalization, which we call query examples $Q_i$.
Our objective is to use the query examples $Q_i$ to enhance the performance of upstream model $\mathcal{M}$ on the unseen task $\mathcal{U}_i$.
For evaluating such unsupervised cross-task generalization methods,  we test the enhanced model with a held-out labeled data of each target task. 

%



\paragraph{Challenges.}
Standard fine-tuning approaches (with or without meta-learning designs) for few-shot cross-task generalization are not feasible here.
We have to adjust the upstream model based on only a few \textit{input-only} examples for the unseen task.
Intuitively, upstream examples that share similar skills with the target task $\mathcal{U}_i$ should be more beneficial than other upstream data.
Thus, one naive idea is to first estimate the {utility} of each upstream example for  $\mathcal{U}_i$ and then re-train a dedicated model $\mathcal{M}_i$ via a weighted learning method (e.g., examples of more utility are trained with larger loss). \relook{\textbf{(updated!)}}

However, such a target-aware weighted re-training method cannot scale, because the upstream data is usually very large and there can be a large number of unseen tasks from users in real-world applications. 
In addition, it is particularly challenging to estimate the utility scores of upstream data for a given unseen task, as we do not have ground-truth annotations for learning this. 
Although there are some existing studies on task-to-task relatedness and transferability~\citep{Vu2020ExploringAP,Lange2021ToSO,Padmakumar2022ExploringTR}, 
most of them are not designed for unsupervised settings and few are done with multi-task (prompted) upstream models.
Moreover, these prior analyses are mainly limited to the task-level analysis and they may not directly generalize to studying example-level utility, which is particularly important for the problem setup of this work.

\section{ReCross: Retrieval Augmentation for Cross-Task Generalization}
\label{sec:methods}



\subsection{Overview}

\begin{figure*}[t]
	\centering
    \vspace{-0.5em}
	\includegraphics[width=0.95\linewidth]{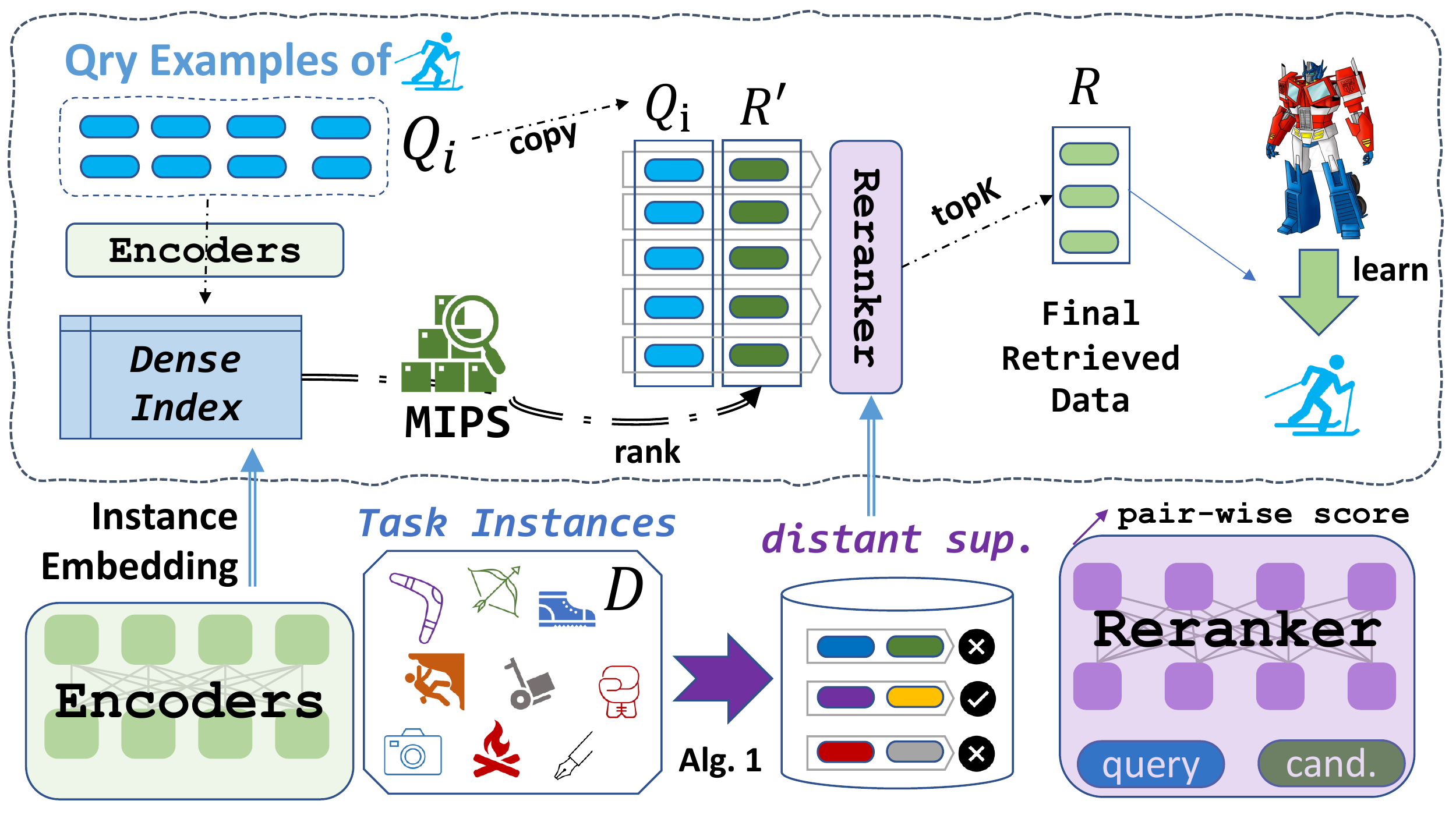} 
	\caption{\textbf{\texttt{ReCross} is a retrieval-augmentation method for unsupervised cross-task generalization.} We reuse the encoder layers of the upstream model (green) to build a dense index, which consists of vectors of the upstream examples $D$. We also propose an algorithm to generate distant supervision for training a reranker, which takes a pair of examples as input and outputs a score.  During the evaluation, we encode query examples $Q_i$ for querying the index to get initial ranking results $R'$, and then pair them with the queries again for reranking. Finally, we take the top-K results (i.e., $R$) for generalizing the upstream model $\mathcal{M}$ to the unseen task $\mathcal{U}_i$. \relook{\textbf{(updated!)}}
	\vspace{-1em}
	}
	\label{fig:pipeline}
\end{figure*}

%


To address the above challenges for unsupervised cross-task generalization,
we propose a retrieval-augmentation method named ReCross.
The ReCross method is also based on the simple idea that we should exploit the upstream examples that enjoy better utility for a given unseen target task. 
Instead of costly re-training from scratch, our method first retrieves a small subset of the upstream data for each unseen task. It then uses them to efficiently fine-tune the upstream model such that the updated model is generalized. 
This can ensure scalability to a great extent and benefit upstream models from re-learning target-specific acquired knowledge for cross-task generalization.

Ideally, we aim to retrieve the upstream examples that are the most beneficial ones for generalizing the upstream model towards a particular unseen task --- ranking the upstream data by their example-level utility.
To achieve this goal while preserving the efficiency, 
we first use the query examples to retrieve initial candidates via efficient maximum inner product search (MIPS) over a dense index, which consists of embedding vectors of all upstream examples (Section~\ref{ssec:dense}).

Based on the candidates from dense retrieval,
we learn a reranking module for further improving the retrieval results (Section~\ref{ssec:rerank}).
The reranker is based on the cross-encoder architecture that takes a query-candidate pair of examples and outputs a more curated score  of utility.
Recall that we do not have any annotation for such example-level utility scores, and the only allowed resources are the upstream data and model.
Therefore, we propose an algorithm to mine distant supervision from the upstream data for learning the reranker  (Section~\ref{ssec:ds}).
The overview of ReCross is shown in Fig.~\ref{fig:pipeline}.

\subsection{Dense Retrieval}
\label{ssec:dense}

To efficiently estimate the example-level utility for generalization,
we propose to first employ a dense retrieval module that ensures high scalability.
Specifically, we build a matrix $\mathbf{D}\in \mathbb{R}^{|D|\times d}$, where each upstream example in ${D}$ is encoded with a dense vector.
Based on this dense index, we can now estimate the utility of an upstream example with its cosine distances to the encoded query examples in $Q$.
That is to say, the upstream examples that are the nearest neighbors of query examples, are more likely to be beneficial for generalizing the upstream model $\mathcal{M}$ to the unseen target task.

To retrieve the candidate set $R'$, we use MIPS to search for the top-$K$ examples for each query example in $Q$, so $K=\ceil{{|R'|}/{|Q|}}$.
(We introduce the details and other aggregation strategies in Appendix.)
This dense-retrieval process is very efficient as we pre-compute the upstream index and perform MIPS for querying the candidates over the index on-the-fly during the generalization stage.
We use the FAISS library~\citep{johnson2019billion} in our implementation.

\paragraph{Instance embeddings.}
The example encoder is a key component of the dense-retrieval pipeline. 
An ideal example encoder is supposed to represent the underlying skills behind an example such that we can use the distances in the result embedding space to estimate utility for cross-task generalization. 
As we do not have annotations of utility scores for training an encoder, 
one may want to use pre-trained sentence embedding models such as SentenceBERT~\citep{reimers2019sentencebertse}.
Our empirical results show that such semantics-based encoders cannot lead to much improvement over random retrieval results.
We think there are two reasons for this failure. 
First, the \textit{semantic} similarities between examples are not suitable for estimating the utility for generalization. 
Second, the \textit{external} encoding modules do not reflect the nature of the upstream model which we want to generalize.

To address these two issues, 
we propose to use the encoding layers of upstream model $\mathcal{M}$ for computing the example embeddings.
Without loss of generality, let us assume $\mathcal{M}$ to be a \textit{text-to-text} Transformer that has multiple layers for both encoders and decoders such as BART.
We encode an example by first obtaining the hidden representation of each token at the last encoder layer (i.e., a sequence of token vectors), and then performing mean-pooling over them to get a single dense vector to represent this example.
By doing this, 
the produced example embeddings reflect the  internal features of the upstream model, which are more relevant to the ``thinking process'' of the upstream model for the examples instead of the shallow semantic information.


\subsection{Reranking Module}
\label{ssec:rerank}
\paragraph{Weakness of the dense retrieval.}
Although dense retrieval is very efficient thanks to the MIPS support, the retrieval performance is limited by its two major weakness.
First, it is a dual-encoder architecture that encodes the candidate example and the query example \textit{separately}, which ignores informative features behind token-to-token attention across a pair of examples.
Second, it is too costly to frequently update the example encoder, which prevents us from learning to refine the retrieval results with distant supervision (if any).
Therefore, we design a re-ranking stage where we train a cross-encoder to further enhance the dense-retrieval results with mined distant supervision (Sec.~\ref{ssec:ds}).


\paragraph{Encoding query-candidate pairs.}
The cross-encoder architecture has been widely used in  sentence-pair classification tasks such as natural language inference and paraphrase detection.
We here use a cross-encoder to encode the \textit{concatenation} of a query example and a candidate example.
Specifically, we fine-tune a RoBERTa~\citep{liu2019roberta} model to classify whether an example pair is a positive or negative match.
The confidence of classifying such a pair to be positive can thus be used as the utility score of the candidate upstream example for this query example. 
On top of this, we then develop a reranking module for further improving retrieval performance as follows.

\paragraph{Scoring paired data.}
To re-rank the initially retrieved data by the dense retriever,
we apply the cross-encoder on all pairs of query examples $Q$ and candidate retrieved examples $R'$, producing scores of all $|Q|*|R|'$ query-candidate pairs.
For each candidate example $r\in R'$, we use the average of all cross-encoder scores involving $r$ as its utility score.
Finally, we take the top-$K$ examples based on this new ranking of candidate examples in $R'$ as the final retrieved data $R$.
We use \textit{upsampling ratio} $\mu$ to denote the ratio between $R'$ and $R$, i.e., $\mu=|R'|/|R|$.


\subsection{Mining Distant Supervision for Reranking}
\label{ssec:ds}
How do we train such a re-ranking module? 
Recall that we only have access to the upstream data $D$ and must not use any data from the unseen tasks at this stage.
Inspired by meta-learning works, we propose an algorithm (Alg.~\ref{alg:ds_reranker}) to mine distant supervision data for creating a \textit{training-as-testing} environment for learning the reranker.  
Our key motivation is to examine the utility scores of candidate examples by assessing the generalization performance of updated models that are fine-tuned with these candidates as if we use them for real unseen tasks.
Such more realistic estimation of utility scores can thus help us train a reranker to predict.

\begin{wrapfigure}[26]{L}{0.5\textwidth}
\vspace{-1.7em}
\begin{minipage}{0.5\textwidth}
\begin{algorithm}[H]
	\small
	\nonl \textbf{Input:}  {{$\mathcal{M}$; $D$; $\mathcal{T}_q$ }} 
	
	\nonl \textbf{Output}: $Z = (Z_{q}, Z_{p}, Z_{n})$  \\ 
	\vspace{-0.5em}
	\hrulefill
	
	
    $D_{\mathcal{T}_q} \xleftarrow{} \{x \in D | x \text{ is an example of } \mathcal{T}_q \} $\\
	$Z_{q} \xleftarrow{}\operatorname{Sample}(D_{\mathcal{T}_q})$;~~
	$H_{q} \xleftarrow{}\operatorname{Sample}(D_{\mathcal{T}_q})$ \\
	$ R_Z \xleftarrow{} \operatorname{DenseRetrieve}(Z_q, D)$ \\
	\comm{\scriptsize Delete retrieved examples from the same task as queries.}\\
	$ R_Z \xleftarrow{} R_Z.\operatorname{discard}(D_{\mathcal{T}_q}) $
	
	\ForEach{ {round}~~ } { 
	 $R_Z.\operatorname{shuffle()}$ \\ 
	 \comm{\scriptsize Split retrieved examples into $n$ groups} \\
	\{$G_1,...,G_n$\} $\xleftarrow{} R_Z.\operatorname{split()}$ \\
    \ForEach{$G_i \text{ in } \{G_1,...,G_n\} $ } { 
        $\mathcal{M}' \xleftarrow{}  \mathcal{M}.\operatorname{copy()}$ \\
        $\mathcal{M}'.\operatorname{fine\_tune}(G_i)$ \\
        $\ell \xleftarrow{} \mathcal{M}'.\operatorname{calc\_loss}(H_q)$ \\
        \ForEach{$x \in G_i$}{
            $scores[x].\operatorname{append}(\ell)$ ~~ \\
            \comm{\scriptsize Score each  in the group w/ the loss.} \\
        }
    }
	}

    \comm{\scriptsize Use mean group score as score for single examples} \\ 
	\ForEach{$x \in R_Z$}{
	    $score[x] \xleftarrow{} \operatorname{mean}(scores[x])$
	}
	\comm{\scriptsize Sort $R_Z$ by $score$ in increasing order.} \\
	
	$R_Z.\operatorname{sort(key:score, order:increasing)}$ \\
	$Z_p \xleftarrow{} $ \text{First $W$ items of $R_Z$} \\
	$Z_n \xleftarrow{} $ \text{Last $W$ items of $R_Z$} \\
	\caption{{\textbf{Distant Supervision Creation }}}\label{alg:ds_reranker}
\end{algorithm}
\end{minipage}
\end{wrapfigure}

Specifically, we define a data point of such distant supervision as a tuple $Z = (Z_{q}, Z_{p}, Z_{n})$:
    1)  $Z_{q}$ is  a set of query examples of a particular task $\mathcal{T}_q$;
    2) $Z_{p}$ is the set of positive examples from other tasks;
    3) $Z_{n}$ is the set of negative examples from other tasks.
We expect that $Z_{p}$ is of more utility for generalization than $Z_{n}$ if $Z_{q}$ would be a query set for the target task $\mathcal{T}_q$.
To this end, we first randomly sample an upstream task $\mathcal{T}_q$ and use a small subset of its training data as the $Z_{q}$. 
Here, we also sample a larger held-out set $H_{q}$ examples of task $\mathcal{T}_q$ to facilitate utility estimation.
Then, we apply the dense retriever using $Z_{q}$ as the query examples and get the retrieval results $R_Z$.
This $R_Z$ is thus the candidate pool where we create $Z_{p}$ and $Z_{n}$. 
That is, $Z_{p} \subset R_Z$ and $Z_{n} \subset R_Z$. 
We discard examples that are from the $\mathcal{T}_q$, so that the generated tuples are closer to the real scenarios where we use the reranker on the query sets of unseen tasks.

Our criteria to select $Z_{p}$ and $Z_{n}$ from $R_Z$ is motivated by the hypothesis that a more suitable set of retrieved examples should improve the performance $\mathcal{M}$ on $\mathcal{T}_i$ after fine-tuning with it.
Therefore, we iteratively sample a small subset from $R_Z$, then fine-tune $\mathcal{M}$ with it, and finally, use the fine-tuned model to evaluate on $Z_{q}'$. 
The performance of such a temporarily fine-tuned model can be seen as the utility score---how well this subset can help generalize $\mathcal{M}$ to the unseen task $\mathcal{T}_q$.
Through multiple rounds of such sample-train-test procedures, we can thus score each example in $R_Z$ by taking the average of all test results where it is involved. 
With such a new ranking of examples in $R_Z$, we take the best $W$ examples as $Z_p$ and the worst $W$ as $Z_n$.


With such distant supervision, we then can create pair of query-positive instances and query-negative instances via pairing $Z_q$ with $Z_p$ and $Z_n$ respectively.
Now we can fine-tune a RoBERTa-base model by concatenating each pair and learning a binary-classification objective.
The output logits of this trained model will be used for the reranking procedure as shown in Sec.~\ref{ssec:rerank}.

\vspace{-0.1cm}
\subsection{Re-learning via Fine-Tuning with Retrieved Data}
\label{ssec:relearn}
\vspace{-0.1cm}
When we have the final retrieved data $R_i$ for a certain query set $Q_i$, we can now enhance the upstream model $\mathcal{M}$ for the unseen task $\mathcal{U}_i$.
We use a small learning rate to continually fine-tune $\mathcal{M}$ with the retrieved upstream examples $R_i$ for a small number of steps. 
We find that the learning rate has to be very small so that this step can be seen as a natural continuation of the finished upstream training and avoid overfitting the retrieved data.
We acknowledge that there could be more effective methods to reuse the query examples $Q$ as guidance for fine-tuning, and we leave this as future work.
Please find more discussion on the hyper-parameter selection and configuration in our appendix. 

\vspace{-0.2cm}
\section{Evaluation}
\label{sec:exps}
\vspace{-0.1cm}

In this section, we first introduce the experimental setups, including the task distribution, upstream learning details, and the configurations of the main experiments.
We present experimental results and reveal some non-trivial findings with extensive analysis that justify the effectiveness of ReCross.


\vspace{-0.2cm}
\subsection{Evaluating Unsupervised Cross-Task Generalization}
\label{ssec:evalmetric}
\vspace{-0.1cm}
We follow~\citet{sanh2021t0} to use the templates from PromptSource~\citep{bach2022promptsource} for converting data of different types of NLP tasks to text-to-text formats.
In total, we have 36 upstream tasks and 10 target unseen tasks for our main experiments.
The upstream tasks are the same as the ones that the T0 models used for upstream learning. 
We follow the evaluation protocol proposed by~\citet{sanh2021t0} and select the target tasks that are significantly different from the upstream tasks.
Besides, we also include 5 additional tasks from the BIG-bench project~\citep{bigbench} to create an even more out-of-distribution set of unseen tasks for analysis. 

\vspace{-0.2cm}

\paragraph{Metric.} When we apply the natural-language templates for the test examples, we only keep the templates that can be evaluated with an exact match (classification, question answering, answer selection, etc.) so that it is feasible to use exact-match for evaluating all tasks. 
To allow a smoother grading, our metric also counts the cases when outputs and truths are sub-strings of each other, which we call \textbf{SoftEM}.   
The only difference between SoftEM and the standard EM is that it also counts the sub-string matches. We observe that sometimes even though T0-like models (including ours) answer the input questions correctly, their raw outputs are not exactly the same as the truth outputs generated by the PromptSource templates.  
In particular, the ground-truth outputs for multiple-choice QA tasks are often in the form of ``[A/B/C/D]: [answer]'', while the models often only output the id of the correct choice (e.g., ``A'') or the text of the answer. 
We also find that the model can output some noise (such as additional punctuation) after the answer (e.g., “True” vs “True.”). The standard EM will discard such matches and cause inaccurate measurements. Although SoftEM might add false positives due to substring matches, we found it is very rare according to our manual inspection of the 10 tasks. 
Therefore, we choose to use SoftEM for a more precise evaluation.
We report the results with the standard EM in Table~\ref{tab:main_EM} that also supports our findings.


\vspace{-0.2cm}
\subsection{BART0: Upstream Learning with a Smaller LM}
\label{ssec:upstreamtraining}
\vspace{-0.1cm}
The T0(pp) models are all very huge, and
the smallest version, T0-3B (3 billion parameters), is still too large to be fine-tuned on popular affordable GPUs. 
We need a parameter-efficient alternative that makes the study on cross-task generalization more accessible to a broader community while keeping the generality.
Thus, we fine-tune a BART-large~\citep{lewis2019bart} (0.4 billion parameters) following the recipe of training T0. 
Specifically, we sample 50k examples at most from each upstream task to build a large upstream dataset consisting of 1.7 million examples (i.e., $|D|=\ $1.7m), 
and then we fine-tune a BART-large with 22k steps with this upstream dataset. 
Finally, we use the fine-tuned checkpoint as our upstream model $\mathcal{M}$ and name it \textbf{BART0}.
Surprisingly, we find that BART0 and T0-3B have comparable zero-shot performance on the unseen target tasks, even though T0-3B is about 8x larger than BART0.
More implementation details  are shown in Appendix.



\vspace{-0.1cm}
\subsection{Setup and Configurations}
\label{ssec:setup_config}
\vspace{-0.1cm}
In our main experiments, 
we use $|Q_i|$ = 16 query examples for each unseen task $\mathcal{U}_i$ and retrieve $|R_i|$ = 512 examples for augmenting BART0.
In the fine-tuning stage, we use a learning rate of 1e-6 and a batch size of 4 to continually fine-tune all layers of BART0 for 2 epochs.
As for re-ranking, we set the upsampling ratio $\mu=2$, meaning that we first retrieve 1024 examples for reranking and use the top 512 ones as the final retrieved data.
To obtain more convincing evaluation results, we average the scores of all target tasks to show the general zero-shot performance. 
For each task $\mathcal{U}_i$, we use five different query sets, $\{Q_i^{(1)}, \dots, Q_i^{(5)}\}$, to conduct \textbf{five individual rounds of retrieval}, thus resulting in five average scores for all tasks.
To get a comprehensive assessment, we report the mean, std, median, min, and max of these five overall scores in the lower part of Table~\ref{tab:main}. 
We present an ablation study on hyper-parameter configurations in Table~\ref{tab:ablation} and include more details in Appendix.

\vspace{-0.1cm}
\subsection{Experimental Results}
\vspace{-0.1cm}

\paragraph{BART0 vs T0-3B.}
As mentioned earlier, we find that BART0 is comparable with the much larger T0-3B in terms of their zero-shot performance on our unseen tasks (41.33 vs 40.38). 
As we use BART0 as our base model for testing different retrieval-augmentation methods, its overall performance \underline{\textit{40.38}} is what we want retrieval-augmentation methods to beat.
Note that when using BART0 and T0-3B for non-retrieval zero-shot inference, they do not use any information from the query examples, so their mean, median, min, and max are always the same.

\begin{table*}[t]
\hspace{-0.8em}
\centering
\scalebox{0.98}{
\begin{tabular}{r|c|c||c|c|c|c||c}
\toprule
Target Task     & T0-3B  & \textbf{\underline{BART0}}   & Random            & SBERT             & ReCross$^\dag$     &\textbf{\underline{ReCross}}   & $\Delta$ \\
\midrule
anli\_r3         & 26.00 & 30.50 & 35.34$_{\pm1.52}$ & 32.64$_{\pm2.53}$ & 36.70$_{\pm0.53}$ & 35.76$_{\pm0.90}$ & 5.26                \\
h-swag        & 34.40 & 39.40 & 33.84$_{\pm5.59}$ & 30.92$_{\pm7.82}$ & 44.36$_{\pm3.07}$ & 47.28$_{\pm2.95}$ & 7.88                \\
cb               & 53.93 & 39.64 & 47.07$_{\pm1.25}$ & 48.00$_{\pm3.28}$ & 44.50$_{\pm4.20}$ & 44.79$_{\pm3.36}$ & 5.15                \\
wic              & 45.70 & 46.70 & 41.04$_{\pm2.18}$ & 46.78$_{\pm2.22}$ & 49.90$_{\pm0.50}$ & 50.58$_{\pm0.24}$ & 3.88                \\
wsc              & 50.00 & 57.88 & 52.50$_{\pm2.29}$ & 52.69$_{\pm6.13}$ & 59.27$_{\pm1.96}$ & 61.46$_{\pm1.47}$ & 3.58                \\
winogrande       & 47.60 & 51.10 & 52.68$_{\pm0.83}$ & 52.18$_{\pm3.20}$ & 54.60$_{\pm1.35}$ & 55.46$_{\pm0.88}$ & 4.36                \\ 
arc-chan.         & 41.30 & 35.70 & 33.28$_{\pm1.50}$ & 37.90$_{\pm1.22}$ & 37.78$_{\pm0.73}$ & 38.44$_{\pm0.99}$ & 2.74                \\
obqa       & 38.50 & 34.40 & 28.72$_{\pm2.46}$ & 33.28$_{\pm1.24}$ & 36.98$_{\pm1.55}$ & 39.58$_{\pm2.80}$ & 5.18                \\
piqa             & 45.30 & 36.10 & 37.00$_{\pm2.71}$ & 38.54$_{\pm2.17}$ & 41.34$_{\pm1.75}$ & 41.42$_{\pm1.02}$ & 5.32                \\
squadv2        & 30.60 & 32.40 & 29.86$_{\pm5.46}$ & 29.46$_{\pm0.84}$ & 30.26$_{\pm1.54}$ & 30.58$_{\pm1.61}$ & -1.82               \\
\midrule \midrule
All@mean   & 41.33 & 40.38 & 39.13$_{\pm2.06}$ & 40.24$_{\pm1.61}$ & 43.57$_{\pm0.68}$ & 44.53$_{\pm0.42}$ & 4.15                \\
@median & 41.33 & 40.38 & 39.93           & 40.91           & 43.43           & 44.31           & 3.93                \\
@min    & 41.33 & 40.38 & 35.66           & 38.28           & 42.65           & 44.16           & 3.77                \\
@max    & 41.33 & 40.38 & 40.59           & 41.76           & 44.51           & 45.07           & 4.69   \\             
\bottomrule

\end{tabular}
}
\caption{\textbf{The main experimental results (\%) for unsupervised cross-task generalization in SoftEM.}
Each result in the upper section is the average (and the std) performance of using 5 different query sets for a task. The lower section of this table reports the mean, max, min, and median of the overall performance (i.e., the average performance on all tasks) of these five rounds.}
\label{tab:main}
\end{table*}

\vspace{-0.2cm}
\paragraph{Random Retrieval.} 
The \textit{Random} column shows the results when we randomly sample $R_i$ from the upstream data $D$ without using any information from $Q_i$. 
\relook{The max performance among the multiple rounds of random retrieval is usually comparable to or larger than the performance of the vanilla BART0 for all tasks. 
For example, although the mean performance of random retrieval for SquadV2 is 29.86, which is smaller than BART0's 32.40, the max of random retrieval is 35.32.  
These maximum performance of ``lucky'' rounds of random retrieval suggest that it is promising to develop better retrieval augmentation methods. Plus, it also suggests that if given suitable retrieved data, such a simple ``re-learning'' method could already enhance the upstream model}.

\vspace{-0.1cm}
\paragraph{SBERT and ReCross$^\dag$.} 
We  use SentenceBERT (SBERT) as a strong baseline method to create a dense index of the upstream data, compared with our proposed indexing method, ReCross$^\dag$ (i.e., ReCross without reranking).
We can see that ReCross$^\dag$ always outperforms the other methods.
Even its minimum performance in the five rounds (\textit{42.65}) is better than the maximum of the SBERT (\textit{41.76)}.
Besides, the standard deviation also becomes much smaller (1.61$\rightarrow$ 0.68), which means that improvement by the ReCross$^\dag$  is more consistent under different query sets. 

The SBERT indexing relies mainly on the \textit{semantic similarities} between a query example and the upstream data.
Instead, our proposed ReCross$^\dag$ uses the hidden representations inside the upstream model $\mathcal{M}$ for representing examples.
We believe using such an indexing method can better help us find examples that share \textit{similar reasoning skills} acquired by the upstream model.

\vspace{-0.1cm}
\paragraph{ReCross = ReCross$^\dag$ + Reranking.}
The full version of our ReCross with reranking can  further improve the performance substantially on multiple dimensions. 
Both all@mean and median are improved by 1 point from the ReCross$^\dag$, and the std is also reduced from 0.68 to 0.42. 
The last column ($\Delta$) in Table~\ref{tab:main} shows its improvement compared to the base model BART0, and we can see that ReCross consistently outperforms non-retrieval methods (e.g., BART0) by a significant gap.

To explore the potential benefits of retrieval-augmentation methods such as our ReCross, we also conduct the same experiments on five tasks selected from the BIG-Bench project. The results are
shown in Table~\ref{tab:bigbench}, where we can see that ReCross still outperforms the non-retrieval methods.
An interesting case is the \texttt{movie\_dialog} task, where the prompt in the template requires a model to output ``same'' or ``different.''
However, both T0-3B and BART0 fail to follow the prompt instruction, and can only output ``yes/no.''
Only when we use retrieval-augmentation methods, there are performance improvement on this task. 

\begin{wraptable}{r}{7cm}
\vspace{-0.4cm}
\begin{minipage}{\textwidth}
\resizebox{0.5\textwidth}{!}{
\begin{tabular}{r|ccc}
\toprule
         Task                                & T0-3B   & BART0  & ReCross            \\ \midrule
hindu\_knowledge                    & 24.75 & 23.48  & 24.87$_{\pm0.27}$  \\
known\_unknowns                      & 47.83 & 43.48 & 47.17$_{\pm1.65}$  \\
logic\_grid\_puzzle                 & 23.60 & 20.70  & 17.12$_{\pm6.29}$  \\
strategyqa                           & 47.70 & 48.30 & 49.76$_{\pm0.80}$  \\
movie\_dialog   & 0.00 & 4.40   & 37.22$_{\pm13.26}$ \\
\midrule
All@Mean& 28.78 & 28.07&		35.23$_{\pm2.85}$  \\
        \bottomrule
\end{tabular}
}
\end{minipage}
\caption{\textbf{Results on a subset of BigBench tasks.}}\label{tab:bigbench}
\vspace{-0.1cm}
\end{wraptable} 




\vspace{-0.1cm}
\subsection{Analysis \& More Findings.}

\vspace{-0.1cm}
\paragraph{More configurations.}
We have used a particular configuration in our main experiments that are in Table~\ref{tab:main}, which is $|Q|$=16, $|R|$=512, and $|u|$=2.
In Table~\ref{tab:ablation}, we explore more configurations as ablation studies.
The ``Main Exp.'' row refers to the results shown in Table~\ref{tab:main}, and the configurations of other rows are only changed with one factor at a time.
Even using a single query example, ReCross is better than BART0.
However, when increasing the query size to 32, we find that the performance starts to decrease, meaning that there could be an optimal query size for a certain $|R|$=512. 
We find that increasing $|R|$ is generally beneficial, while the all@mean decreases when $|R|$ is changed from 512 to 1024, although the max and the median slightly increased.
Finally, we see that increasing $\mu$ increases the std. and does not improve the overall performance.



\begin{wraptable}{r}{7cm}
\vspace{-0.2cm}
\begin{minipage}{\textwidth}
\resizebox{0.5\textwidth}{!}{
\begin{tabular}{c|ccccc}
\toprule
Setup\textbackslash All@        & Mean & std.    & Min     & Max     & Median  \\
           \midrule
\rowcolor{gray!10} Main Exp.   & 44.53  & 0.42  & 44.16  & 45.07  & 44.31  \\ \midrule
$|Q|$=1~~    & 43.20  & 0.83  & 42.58  & 44.58  & 42.88  \\
$|Q|$=8~~    & 43.67  & 0.90  & 42.09  & 44.32  & 43.90  \\
$|Q|$=32   & 42.52  & 1.17  & 40.52  & 43.40  & 42.96  \\\midrule
$|R|$=256~~  & 40.80  & 0.83  & 39.45  & 41.68  & 40.96  \\
$|R|$=1024 & 44.02  & 1.43  & 42.26  & 45.35  & 44.59  \\\midrule
$\mu$=3    & 43.92  & 0.58  & 43.08  & 44.57  & 43.89  \\

$\mu$=4  & 43.91 &	0.99 &	42.76 &	45.10 &	44.26  \\
        \bottomrule
\end{tabular}
}
\end{minipage}
\caption{\textbf{The ablation study of ReCross.}}\label{tab:ablation}
\end{wraptable}

\vspace{-0.1cm}
\paragraph{Retrieved data distribution.}

\begin{figure*}[t]
    \vspace{-1em}
	\hspace{-1.5em}
	\includegraphics[width=1.05\linewidth]{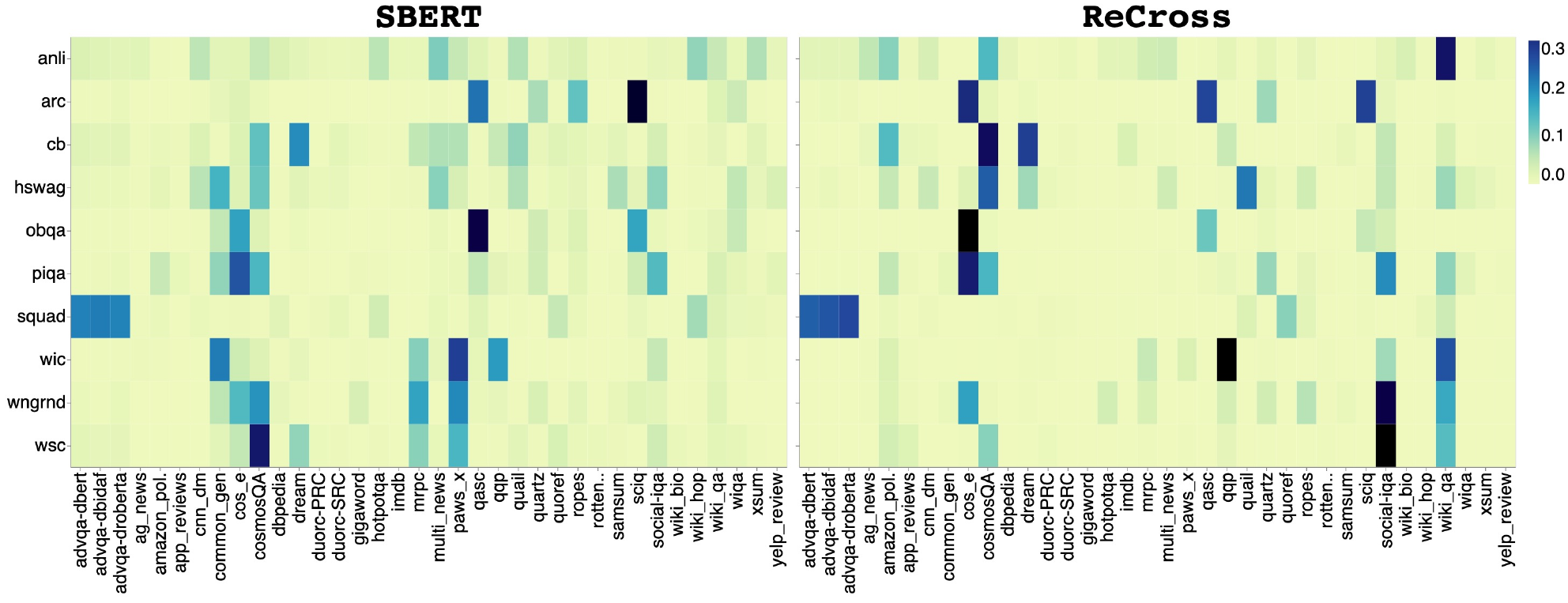} 
	\caption{\textbf{The mapping between unseen tasks (as rows) and upstream tasks (as columns).} The darker upstream tasks take more percentage in retrieved data. For example, for the task \texttt{WIC}, ReCross retrieves a plurality of examples from \texttt{QQP} (about 30\% of the retrieved examples).  \vspace{-1.5em}
	}
	\label{fig:corr}
\end{figure*}

Figure~\ref{fig:corr} presents the difference between the methods in terms of their retrieved data.
We draw the distribution of the retrieved data among different upstream tasks for each unseen task individually.
From the heatmap, we can see that
ReCross tends to have more dominant retrieved tasks (i.e., darker cells), while SBERT's results are more sparse.
They both can identify that \texttt{squad} is most similar to the \texttt{adversarial\_qa} tasks.
Their behaviors are very different too.
Taking the unseen task \texttt{winogrande} (wngrnd) as an example, 
we can see that the SBERT retrieves from multiple upstream tasks such as \texttt{paws-x} and \texttt{cosmosQA} , but the ReCross mainly retrieves from \texttt{social-iqa}, \texttt{wiki-qa}, and \texttt{cos-e}.
The experimental results in Table~\ref{tab:main} show that ReCross produces a better performance than SBERT (i.e., 55.46 vs 52.18),
while it is not clear how we can predict such task correlation in advance.
This suggests that we should explore more about the utility of instances and tasks in future work.


\paragraph{More analysis.}
In the appendix, we further presented some analysis to help understand “how” and “when” the retrieval augmentation works: Table~\ref{tab:remove}, Table~\ref{tab:temp_perturb}, Appendix A.1~A.2, and Appendix B.
We investigate whether the utility of upstream examples in retrieval augmentation is related to the similarity in terms of the task formats. From Appendix A.1, we found some counterintuitive results. For example, if removing MCQA upstream tasks from the upstream index, then the ARC target task can have an even better performance, although it is an MCQA-formatted task. Thus, we hypothesize that similarity in terms of reasoning types is more important than format similarity for retrieval augmentation. After all, the upstream model has been already trained to work with these basic task formats. Re-learning the tasks of the same format might lead the model to overfit the seen domains. Additionally, to provide a more concrete analysis, we also present case studies with two specific tasks (CB and SQUADv2) in Appendix~\ref{app:case}.

Moreover, we conjecture the natural language instructions in the templates are necessary for ReCross to get impressive results. Therefore, we investigated two ways of perturbing the instructions and monitoring the performance changes in Appendix A.2. We find it is indeed true that perturbations of the instructions will lead to much worse performance.
We believe that a rigorous, principled way of analyzing the correlation between query and retrieval examples will be a great future direction, given the strong evidence that ReCross works so well as such a simple method.

\section{More Discussion}

\subsection{Practicality of unsupervised setting.}
\paragraph{Cost of obtaining task labels}
The unsupervised setting in the paper does not require any human annotation of labels. For some tasks (NLG tasks in particular, e.g., summarization), the expected output (label) are open-ended and possibly lengthy and thus human annotation is much more expensive and time-consuming. Also, few-shot learning must ask humans to label examples for each new task, and it is thus less practical when there are a large number of emerging tasks from the users. Meanwhile, ReCross requires only a natural-language task template, which does not require potentially expensive manual annotation or domain expertise.

\paragraph{Scalability \& Real-Time response}
Deploying the ReCross pipeline is a one-time process. All we need to do is to pre-compute the upstream index with LM and configure the reranker (a simple masked LM) by running our script. In production, once the users input the examples with NL instructions, we do not need to wait for any human annotations anymore, so it is much more efficient in the long run.
In the scenarios where users only provide one query example and want to get its label from the model, ReCross also shows great performance (i.e., |Q|=1 in Table~\ref{tab:main}). 
It is then impractical to assume there are a few labeled data from the users too in such cases.

\subsection{Empirical studies}
The unsupervised ReCross performance is comparable to few-shot learning with label annotations. In Appendix D.2, we report the performance of directly fine-tuning BART0 with the labeled query examples. Although it is an unfair comparison with our previous ReCross results, we found that they are comparable. 
More importantly, the ReCross framework does not conflict with the few-shot setting. Given a labeled query set for a target task, retrieved examples from the ReCross can still improve few-shot learning as additional training data. We designed two simple methods for applying ReCross under the few-shot setting and report the empirical results in Appendix D.2. It turns out that ReCross can also boost the performance under the few-shot setting by about 3 points.

\vspace{-0.2cm}
\section{Related Work}
\label{sec:related}

\vspace{-0.1cm}
\paragraph{Multi-task training for task generalization.}
Text-to-text Transformer language models such as T5 enable us to train a multi-task NLP model with a more straightforward recipe: mixing the data of multiple tasks into a unified seq2seq format, and then fine-tuning text-to-text LMs for implicit multi-task learning. 
UnifiedQA~\citep{khashabi2020unifiedqa} is among the first works in this direction.
Although it shows great generalization performance within QA tasks, it can hardly generalize to other NLP tasks.  
Recent works, such as CrossFit~\citep{ye-etal-2021-crossfit}, ExT5~\citep{aribandi2022ext}, FLAN~\citep{Wei2021FinetunedLM},  T0~\citep{sanh2021t0}, and InstructGPT~\cite{InstructGPT} focus on how to generalize a massively multi-task model across task boundaries in a much broader context.

Particularly, in the CrossFit framework~\citep{ye-etal-2021-crossfit},
cross-task generalization requires a small number of labeled instances of the target task for fine-tuning.
It is because the templates of CrossFit use the task names as the hard prefixes. 
Therefore, it is necessary to fine-tune the upstream model with a few examples that have the target task names as prefixes (i.e., few-shot learning), but this largely limits the application scenarios of these multi-task NLP models in practice. 
We instead focus on \textit{unsupervised} cross-task generalization, where there is no \textit{labeled} data of an unseen task (i.e., zero-shot learning). 
Using natural-language instructions as prompts, both FLAN and T0 show that it is promising to perform \textit{zero-shot} cross-task generalization.

In this work, we also focus on such an unsupervised setting for cross-task generalization, while our problem setup is a bit different from the ones used in T0 and FLAN.
As for the assumption about the unlabeled data, their setups can be seen as a special case of ours when $|Q|=1$ for all unseen tasks.
The evaluation protocols of T0 and FLAN assess the generalization performance of the upstream model as it is, and thus their evaluation is more about the quality of templates and the upstream training tricks.
In contrast, our evaluation protocol can also study how to efficiently adjust the upstream model such that the updated models can generalize to new tasks without labeled data.
Thus, we believe ours is a more general setup for studying unsupervised cross-task generalization.


\vspace{-0.2cm}
\paragraph{Retrieval augmentation in NLP.}
We aim to retrieve useful examples from the upstream data and re-learning them for cross-task generalization.
The proposed ReCross pipeline is inspired by open-ended QA methods such as DPR~\citep{dpr},  DrFact~\citep{Lin2021DifferentiableOC}, and RAG~\citep{Lewis2020RetrievalAugmentedGF}.
Retrieval augmentation also shows great performance in pre-training LMs~\citep{Guu2020REALMRL}.
Besides, \cite{Wang2022TrainingDI} shows that learning with similar data via retrieval augmentation can improve the performance of a task-specific model.
\cite{Rubin2021LearningTR} show that retrieving better demonstration examples is also helpful for in-context few-shot learning of GPT-3 style language models~\citep{brown2020language}.
The key challenge in the problem setup of this work is to predict the utility of the examples for unseen tasks with the consideration of efficiency and scalability.
We have discussed more details about this challenge and related works in Sec.~\ref{sec:problem}.


\vspace{-0.1cm}
\section{Conclusion \& Future Directions}
\label{sec:conclusion}  
\vspace{-0.2cm}
We demonstrate that retrieval augmentation can largely improve the cross-task generalization ability to multitask LMs in unsupervised settings.
Our proposed method, ReCross, is a straightforward yet effective retrieval method that combines both efficient dense retrieval and effective pair-wise reranking.
Our empirical results show that it significantly outperforms both non-retrieval methods and other baseline methods.
We perform ablation studies showing the impact of changing query sizes, retrieval sizes, upsampling ratios, etc.
We also find the distribution of retrieved data for analyzing the behavior differences between ReCross and others.
We believe that our paper will spur further research on retrieval-augmentation methods for cross-task generalization.
Interesting future directions include: 1) improve the re-learning stage by including more information from query examples, 2) extend the distant supervision mining process as a self-training procedure, 3) rigorously analyze the correlation between upstream data and target tasks, etc.



\section*{Acknowledgments}
{\small
This research is supported in part by the Office of the Director of National Intelligence (ODNI), Intelligence Advanced Research Projects Activity (IARPA), via Contract No. 2019-19051600007, the DARPA MCS program under Contract No. N660011924033, the Defense Advanced Research Projects Agency with awards W911NF-19-20271, NSF IIS 2048211, and gift awards from Google, Amazon, JP Morgan, and Sony. We thank all collaborators in USC and the NeurIPS 2022 reviewers for their constructive feedback on the work. 
}
\bibliography{custom}

\begin{thebibliography}{23}
\expandafter\ifx\csname natexlab\endcsname\relax\def\natexlab#1{#1}\fi

\bibitem[{Aribandi et~al.(2022)Aribandi, Tay, Schuster, Rao, Zheng, Mehta,
  Zhuang, Tran, Bahri, Ni, Gupta, Hui, Ruder, and Metzler}]{aribandi2022ext}
Vamsi Aribandi, Yi~Tay, Tal Schuster, Jinfeng Rao, Huaixiu~Steven Zheng,
  Sanket~Vaibhav Mehta, Honglei Zhuang, Vinh~Q. Tran, Dara Bahri, Jianmo Ni,
  Jai Gupta, Kai Hui, Sebastian Ruder, and Donald Metzler. 2022.
\newblock \href {https://openreview.net/forum?id=Vzh1BFUCiIX} {Ext5: Towards
  extreme multi-task scaling for transfer learning}.
\newblock In \emph{International Conference on Learning Representations}.

\bibitem[{Bach et~al.(2022)Bach, Sanh, Yong, Webson, Raffel, Nayak, Sharma,
  Kim, Bari, Fevry, Alyafeai, Dey, Santilli, Sun, Ben-david, Xu, Chhablani,
  Wang, Fries, Al-shaibani, Sharma, Thakker, Almubarak, Tang, Radev, Jiang, and
  Rush}]{bach2022promptsource}
Stephen Bach, Victor Sanh, Zheng~Xin Yong, Albert Webson, Colin Raffel,
  Nihal~V. Nayak, Abheesht Sharma, Taewoon Kim, M~Saiful Bari, Thibault Fevry,
  Zaid Alyafeai, Manan Dey, Andrea Santilli, Zhiqing Sun, Srulik Ben-david,
  Canwen Xu, Gunjan Chhablani, Han Wang, Jason Fries, Maged Al-shaibani, Shanya
  Sharma, Urmish Thakker, Khalid Almubarak, Xiangru Tang, Dragomir Radev, Mike
  Tian-jian Jiang, and Alexander Rush. 2022.
\newblock \href {https://doi.org/10.18653/v1/2022.acl-demo.9}
  {{P}rompt{S}ource: An integrated development environment and repository for
  natural language prompts}.
\newblock In \emph{Proceedings of the 60th Annual Meeting of the Association
  for Computational Linguistics: System Demonstrations}, pages 93--104, Dublin,
  Ireland. Association for Computational Linguistics.

\bibitem[{Brown et~al.(2020)Brown, Mann, Ryder, Subbiah, Kaplan, Dhariwal,
  Neelakantan, Shyam, Sastry, Askell, Agarwal, Herbert{-}Voss, Krueger,
  Henighan, Child, Ramesh, Ziegler, Wu, Winter, Hesse, Chen, Sigler, Litwin,
  Gray, Chess, Clark, Berner, McCandlish, Radford, Sutskever, and
  Amodei}]{brown2020language}
Tom~B. Brown, Benjamin Mann, Nick Ryder, Melanie Subbiah, Jared Kaplan,
  Prafulla Dhariwal, Arvind Neelakantan, Pranav Shyam, Girish Sastry, Amanda
  Askell, Sandhini Agarwal, Ariel Herbert{-}Voss, Gretchen Krueger, Tom
  Henighan, Rewon Child, Aditya Ramesh, Daniel~M. Ziegler, Jeffrey Wu, Clemens
  Winter, Christopher Hesse, Mark Chen, Eric Sigler, Mateusz Litwin, Scott
  Gray, Benjamin Chess, Jack Clark, Christopher Berner, Sam McCandlish, Alec
  Radford, Ilya Sutskever, and Dario Amodei. 2020.
\newblock \href
  {https://proceedings.neurips.cc/paper/2020/hash/1457c0d6bfcb4967418bfb8ac142f64a-Abstract.html}
  {Language models are few-shot learners}.
\newblock In \emph{Advances in Neural Information Processing Systems 33: Annual
  Conference on Neural Information Processing Systems 2020, NeurIPS 2020,
  December 6-12, 2020, virtual}.

\bibitem[{Guu et~al.(2020)Guu, Lee, Tung, Pasupat, and Chang}]{Guu2020REALMRL}
Kelvin Guu, Kenton Lee, Zora Tung, Panupong Pasupat, and Ming-Wei Chang. 2020.
\newblock \href {https://arxiv.org/abs/2002.08909} {Retrieval augmented
  language model pre-training}.

\bibitem[{Johnson et~al.(2019)Johnson, Douze, and
  J{\'e}gou}]{johnson2019billion}
Jeff Johnson, Matthijs Douze, and Herv{\'e} J{\'e}gou. 2019.
\newblock Billion-scale similarity search with gpus.
\newblock \emph{IEEE Transactions on Big Data}.

\bibitem[{Karpukhin et~al.(2020)Karpukhin, Oguz, Min, Lewis, Wu, Edunov, Chen,
  and Yih}]{dpr}
Vladimir Karpukhin, Barlas Oguz, Sewon Min, Patrick Lewis, Ledell Wu, Sergey
  Edunov, Danqi Chen, and Wen-tau Yih. 2020.
\newblock \href {https://doi.org/10.18653/v1/2020.emnlp-main.550} {Dense
  passage retrieval for open-domain question answering}.
\newblock In \emph{Proceedings of the 2020 Conference on Empirical Methods in
  Natural Language Processing (EMNLP)}, pages 6769--6781, Online. Association
  for Computational Linguistics.

\bibitem[{Khashabi et~al.(2020)Khashabi, Min, Khot, Sabharwal, Tafjord, Clark,
  and Hajishirzi}]{khashabi2020unifiedqa}
Daniel Khashabi, Sewon Min, Tushar Khot, Ashish Sabharwal, Oyvind Tafjord,
  Peter Clark, and Hannaneh Hajishirzi. 2020.
\newblock \href {https://doi.org/10.18653/v1/2020.findings-emnlp.171}
  {{UNIFIEDQA}: Crossing format boundaries with a single {QA} system}.
\newblock In \emph{Findings of the Association for Computational Linguistics:
  EMNLP 2020}, pages 1896--1907, Online. Association for Computational
  Linguistics.

\bibitem[{Lange et~al.(2021)Lange, Str{\"o}tgen, Adel, and
  Klakow}]{Lange2021ToSO}
Lukas Lange, Jannik Str{\"o}tgen, Heike Adel, and Dietrich Klakow. 2021.
\newblock \href {https://doi.org/10.18653/v1/2021.emnlp-main.689} {To share or
  not to share: {P}redicting sets of sources for model transfer learning}.
\newblock In \emph{Proceedings of the 2021 Conference on Empirical Methods in
  Natural Language Processing}, pages 8744--8753, Online and Punta Cana,
  Dominican Republic. Association for Computational Linguistics.

\bibitem[{Lewis et~al.(2020{\natexlab{a}})Lewis, Liu, Goyal, Ghazvininejad,
  Mohamed, Levy, Stoyanov, and Zettlemoyer}]{lewis2019bart}
Mike Lewis, Yinhan Liu, Naman Goyal, Marjan Ghazvininejad, Abdelrahman Mohamed,
  Omer Levy, Veselin Stoyanov, and Luke Zettlemoyer. 2020{\natexlab{a}}.
\newblock \href {https://doi.org/10.18653/v1/2020.acl-main.703} {{BART}:
  Denoising sequence-to-sequence pre-training for natural language generation,
  translation, and comprehension}.
\newblock In \emph{Proceedings of the 58th Annual Meeting of the Association
  for Computational Linguistics}, pages 7871--7880, Online. Association for
  Computational Linguistics.

\bibitem[{Lewis et~al.(2020{\natexlab{b}})Lewis, Perez, Piktus, Petroni,
  Karpukhin, Goyal, K{\"{u}}ttler, Lewis, Yih, Rockt{\"{a}}schel, Riedel, and
  Kiela}]{Lewis2020RetrievalAugmentedGF}
Patrick S.~H. Lewis, Ethan Perez, Aleksandra Piktus, Fabio Petroni, Vladimir
  Karpukhin, Naman Goyal, Heinrich K{\"{u}}ttler, Mike Lewis, Wen{-}tau Yih,
  Tim Rockt{\"{a}}schel, Sebastian Riedel, and Douwe Kiela. 2020{\natexlab{b}}.
\newblock \href
  {https://proceedings.neurips.cc/paper/2020/hash/6b493230205f780e1bc26945df7481e5-Abstract.html}
  {Retrieval-augmented generation for knowledge-intensive {NLP} tasks}.
\newblock In \emph{Advances in Neural Information Processing Systems 33: Annual
  Conference on Neural Information Processing Systems 2020, NeurIPS 2020,
  December 6-12, 2020, virtual}.

\bibitem[{Lin et~al.(2021)Lin, Sun, Dhingra, Zaheer, Ren, and
  Cohen}]{Lin2021DifferentiableOC}
Bill~Yuchen Lin, Haitian Sun, Bhuwan Dhingra, Manzil Zaheer, Xiang Ren, and
  William Cohen. 2021.
\newblock \href {https://doi.org/10.18653/v1/2021.naacl-main.366}
  {Differentiable open-ended commonsense reasoning}.
\newblock In \emph{Proceedings of the 2021 Conference of the North American
  Chapter of the Association for Computational Linguistics: Human Language
  Technologies}, pages 4611--4625, Online. Association for Computational
  Linguistics.

\bibitem[{Liu et~al.(2019)Liu, Ott, Goyal, Du, Joshi, Chen, Levy, Lewis,
  Zettlemoyer, and Stoyanov}]{liu2019roberta}
Yinhan Liu, Myle Ott, Naman Goyal, Jingfei Du, Mandar Joshi, Danqi Chen, Omer
  Levy, Mike Lewis, Luke Zettlemoyer, and Veselin Stoyanov. 2019.
\newblock \href {https://arxiv.org/abs/1907.11692} {Roberta: A robustly
  optimized bert pretraining approach}.
\newblock \emph{ArXiv preprint}, abs/1907.11692.

\bibitem[{Ouyang et~al.(2022)Ouyang, Wu, Jiang, Almeida, Wainwright, Mishkin,
  Zhang, Agarwal, Slama, Ray, Schulman, Hilton, Kelton, Miller, Simens, Askell,
  Welinder, Christiano, Leike, and Lowe}]{InstructGPT}
Long Ouyang, Jeff Wu, Xu~Jiang, Diogo Almeida, Carroll~L. Wainwright, Pamela
  Mishkin, Chong Zhang, Sandhini Agarwal, Katarina Slama, Alex Ray, John
  Schulman, Jacob Hilton, Fraser Kelton, Luke~E. Miller, Maddie Simens, Amanda
  Askell, Peter Welinder, Paul~Francis Christiano, Jan Leike, and Ryan~J. Lowe.
  2022.
\newblock Training language models to follow instructions with human feedback.
\newblock \emph{ArXiv}, abs/2203.02155.

\bibitem[{Padmakumar et~al.(2022)Padmakumar, Lausen, Ballesteros, Zha, He, and
  Karypis}]{Padmakumar2022ExploringTR}
Vishakh Padmakumar, Leonard Lausen, Miguel Ballesteros, Sheng Zha, He~He, and
  George Karypis. 2022.
\newblock \href {https://doi.org/10.18653/v1/2022.naacl-main.183} {Exploring
  the role of task transferability in large-scale multi-task learning}.
\newblock In \emph{Proceedings of the 2022 Conference of the North American
  Chapter of the Association for Computational Linguistics: Human Language
  Technologies}, pages 2542--2550, Seattle, United States. Association for
  Computational Linguistics.

\bibitem[{Raffel et~al.(2020)Raffel, Shazeer, Roberts, Lee, Narang, Matena,
  Zhou, Li, and Liu}]{t5}
Colin Raffel, Noam Shazeer, Adam Roberts, Katherine Lee, Sharan Narang, Michael
  Matena, Yanqi Zhou, Wei Li, and Peter~J Liu. 2020.
\newblock Exploring the limits of transfer learning with a unified text-to-text
  transformer.
\newblock \emph{Journal of Machine Learning Research}, 21(140):1--67.

\bibitem[{Reimers and Gurevych(2019)}]{reimers2019sentencebertse}
Nils Reimers and Iryna Gurevych. 2019.
\newblock \href {https://doi.org/10.18653/v1/D19-1410} {Sentence-{BERT}:
  Sentence embeddings using {S}iamese {BERT}-networks}.
\newblock In \emph{Proceedings of the 2019 Conference on Empirical Methods in
  Natural Language Processing and the 9th International Joint Conference on
  Natural Language Processing (EMNLP-IJCNLP)}, pages 3982--3992, Hong Kong,
  China. Association for Computational Linguistics.

\bibitem[{Rubin et~al.(2022)Rubin, Herzig, and Berant}]{Rubin2021LearningTR}
Ohad Rubin, Jonathan Herzig, and Jonathan Berant. 2022.
\newblock \href {https://doi.org/10.18653/v1/2022.naacl-main.191} {Learning to
  retrieve prompts for in-context learning}.
\newblock In \emph{Proceedings of the 2022 Conference of the North American
  Chapter of the Association for Computational Linguistics: Human Language
  Technologies}, pages 2655--2671, Seattle, United States. Association for
  Computational Linguistics.

\bibitem[{Sanh et~al.(2021)Sanh, Webson, Raffel, Bach, Sutawika, Alyafeai,
  Chaffin, Stiegler, Scao, Raja, Dey, Bari, Xu, Thakker, Sharma, Szczechla,
  Kim, Chhablani, Nayak, Datta, Chang, Jiang, Wang, Manica, Shen, Yong, Pandey,
  Bawden, Wang, Neeraj, Rozen, Sharma, Santilli, Fevry, Fries, Teehan,
  Biderman, Gao, Bers, Wolf, and Rush}]{sanh2021t0}
Victor Sanh, Albert Webson, Colin Raffel, Stephen~H. Bach, Lintang Sutawika,
  Zaid Alyafeai, Antoine Chaffin, Arnaud Stiegler, Teven~Le Scao, Arun Raja,
  Manan Dey, M~Saiful Bari, Canwen Xu, Urmish Thakker, Shanya~Sharma Sharma,
  Eliza Szczechla, Taewoon Kim, Gunjan Chhablani, Nihal Nayak, Debajyoti Datta,
  Jonathan Chang, Mike Tian-Jian Jiang, Han Wang, Matteo Manica, Sheng Shen,
  Zheng~Xin Yong, Harshit Pandey, Rachel Bawden, Thomas Wang, Trishala Neeraj,
  Jos Rozen, Abheesht Sharma, Andrea Santilli, Thibault Fevry, Jason~Alan
  Fries, Ryan Teehan, Stella Biderman, Leo Gao, Tali Bers, Thomas Wolf, and
  Alexander~M. Rush. 2021.
\newblock \href {http://arxiv.org/abs/2110.08207} {Multitask prompted training
  enables zero-shot task generalization}.

\bibitem[{Srivastava et~al.(2022)Srivastava, Rastogi, Rao, Shoeb, Abid, Fisch,
  Brown, Santoro, Gupta, Garriga-Alonso, Kluska, Lewkowycz, Agarwal, Power,
  Ray, Warstadt, Kocurek, Safaya, Tazarv, Xiang, Parrish, Nie, Hussain, Askell,
  Dsouza, Rahane, Iyer, Andreassen, Santilli, Stuhlmuller, Dai, La, Lampinen,
  Zou, Jiang, Chen, Vuong, Gupta, Gottardi, Norelli, Venkatesh, Gholamidavoodi,
  Tabassum, Menezes, Kirubarajan, Mullokandov, Sabharwal, Herrick, Efrat,
  Erdem, Karakacs, Roberts, Loe, Zoph, Bojanowski, Ozyurt, Hedayatnia,
  Neyshabur, Inden, Stein, Ekmekci, Lin, Howald, Diao, Dour, Stinson, Argueta,
  Ram'irez, Singh, Rathkopf, Meng, Baral, Wu, Callison-Burch, Waites, Voigt,
  Manning, Potts, Ramirez, Rivera, Siro, Raffel, Ashcraft, Garbacea, Sileo,
  Garrette, Hendrycks, Kilman, Roth, Freeman, Khashabi, Levy, Gonz'alez,
  Hernandez, Chen, Ippolito, Gilboa, Dohan, Drakard, Jurgens, Datta, Ganguli,
  Emelin, Kleyko, Yuret, Chen, Tam, Hupkes, Misra, Buzan, Mollo, Yang, Lee,
  Shutova, Cubuk, Segal, Hagerman, Barnes, Donoway, Pavlick, Rodol{\`a}, Lam,
  Chu, Tang, Erdem, Chang, Chi, Dyer, Jerzak, Kim, Manyasi, Zheltonozhskii,
  Xia, Siar, Mart'inez-Plumed, Happ'e, Chollet, Rong, Mishra, Winata, de~Melo,
  Kruszewski, Parascandolo, Mariani, Wang, Jaimovitch-L'opez, Betz, Gur-Ari,
  Galijasevic, Kim, Rashkin, Hajishirzi, Mehta, Bogar, Shevlin, Sch{\"u}tze,
  Yakura, Zhang, Wong, Ng, Noble, Jumelet, Geissinger, Kernion, Hilton, Lee,
  Fisac, Simon, Koppel, Zheng, Zou, Koco'n, Thompson, Kaplan, Radom,
  Sohl-Dickstein, Phang, Wei, Yosinski, Novikova, Bosscher, Marsh, Kim, Taal,
  Engel, Alabi, Xu, Song, Tang, Waweru, Burden, Miller, Balis, Berant,
  Frohberg, Rozen, Hern{\'a}ndez-Orallo, Boudeman, Jones, Tenenbaum, Rule,
  Chua, Kanclerz, Livescu, Krauth, Gopalakrishnan, Ignatyeva, Markert, Dhole,
  Gimpel, Omondi, Mathewson, Chiafullo, Shkaruta, Shridhar, McDonell,
  Richardson, Reynolds, Gao, Zhang, Dugan, Qin, Contreras-Ochando, Morency,
  Moschella, Lam, Noble, Schmidt, He, Col'on, Metz, cSenel, Bosma, Sap, ter
  Hoeve, Andrea, Farooqi, Faruqui, Mazeika, Baturan, Marelli, Maru, Quintana,
  Tolkiehn, Giulianelli, Lewis, Potthast, Leavitt, Hagen, Schubert,
  Baitemirova, Arnaud, McElrath, Yee, Cohen, Gu, Ivanitskiy, Starritt, Strube,
  Swkedrowski, Bevilacqua, Yasunaga, Kale, Cain, Xu, Suzgun, Tiwari, Bansal,
  Aminnaseri, Geva, Gheini, MukundVarma, Peng, Chi, Lee, Krakover, Cameron,
  Roberts, Doiron, Nangia, Deckers, Muennighoff, Keskar, Iyer, Constant,
  Fiedel, Wen, Zhang, Agha, Elbaghdadi, Levy, Evans, Casares, Doshi, Fung,
  Liang, Vicol, Alipoormolabashi, Liao, Liang, Chang, Eckersley, Htut, Hwang,
  Milkowski, Patil, Pezeshkpour, Oli, Mei, LYU, Chen, Banjade, Rudolph,
  Gabriel, Habacker, Delgado, Milli{\`e}re, Garg, Barnes, Saurous, Arakawa,
  Raymaekers, Frank, Sikand, Novak, Sitelew, Bras, Liu, Jacobs, Zhang,
  Salakhutdinov, Chi, Lee, Stovall, Teehan, Yang, Singh, Mohammad, Anand,
  Dillavou, Shleifer, Wiseman, Gruetter, Bowman, Schoenholz, Han, Kwatra, Rous,
  Ghazarian, Ghosh, Casey, Bischoff, Gehrmann, Schuster, Sadeghi, Hamdan, Zhou,
  Srivastava, Shi, Singh, Asaadi, Gu, Pachchigar, Toshniwal, Upadhyay, Debnath,
  Shakeri, Thormeyer, Melzi, Reddy, Makini, hwan Lee, Torene, Hatwar, Dehaene,
  Divic, Ermon, Biderman, Lin, Prasad, Piantadosi, Shieber, Misherghi,
  Kiritchenko, Mishra, Linzen, Schuster, Li, Yu, Ali, Hashimoto, Wu, Desbordes,
  Rothschild, Phan, Wang, Nkinyili, Schick, Kornev, Telleen-Lawton, Tunduny,
  Gerstenberg, Chang, Neeraj, Khot, Shultz, Shaham, Misra, Demberg, Nyamai,
  Raunak, Ramasesh, Prabhu, Padmakumar, Srikumar, Fedus, Saunders, Zhang,
  Vossen, Ren, Tong, Wu, Shen, Yaghoobzadeh, Lakretz, Song, Bahri, Choi, Yang,
  Hao, Chen, Belinkov, Hou, Hou, Bai, Seid, Xinran, Zhao, Wang, Wang, Wang, Wu,
  Singh, and Shaham}]{bigbench}
Aarohi Srivastava, Abhinav Rastogi, Abhishek~B Rao, Abu Awal~Md Shoeb, Abubakar
  Abid, Adam Fisch, Adam~R. Brown, Adam Santoro, Aditya Gupta, Adri{\`a}
  Garriga-Alonso, Agnieszka Kluska, Aitor Lewkowycz, Akshat Agarwal, Alethea
  Power, Alex Ray, Alex Warstadt, Alexander~W. Kocurek, Ali Safaya, Ali Tazarv,
  Alice Xiang, Alicia Parrish, Allen Nie, Aman Hussain, Amanda Askell, Amanda
  Dsouza, Ameet~Annasaheb Rahane, Anantharaman~S. Iyer, Anders~Johan
  Andreassen, Andrea Santilli, Andreas Stuhlmuller, Andrew~M. Dai, Andrew~D.
  La, Andrew~Kyle Lampinen, Andy Zou, Angela Jiang, Angelica Chen, Anh Vuong,
  Animesh Gupta, Anna Gottardi, Antonio Norelli, Anu Venkatesh, Arash
  Gholamidavoodi, Arfa Tabassum, Arul Menezes, Arun Kirubarajan, Asher
  Mullokandov, Ashish Sabharwal, Austin Herrick, Avia Efrat, Aykut Erdem, Ayla
  Karakacs, Bridget~R. Roberts, Bao~Sheng Loe, Barret Zoph, Bartlomiej
  Bojanowski, Batuhan Ozyurt, Behnam Hedayatnia, Behnam Neyshabur, Benjamin
  Inden, Benno Stein, Berk Ekmekci, Bill~Yuchen Lin, Blake~Stephen Howald,
  Cameron Diao, Cameron Dour, Catherine Stinson, Cedrick Argueta, C'esar~Ferri
  Ram'irez, Chandan Singh, Charles Rathkopf, Chenlin Meng, Chitta Baral, Chiyu
  Wu, Chris Callison-Burch, Chris Waites, Christian Voigt, Christopher~D.
  Manning, Christopher Potts, Cindy~Tatiana Ramirez, Clara Rivera, Clemencia
  Siro, Colin Raffel, Courtney Ashcraft, Cristina Garbacea, Damien Sileo,
  Daniel~H Garrette, Dan Hendrycks, Dan Kilman, Dan Roth, Daniel Freeman,
  Daniel Khashabi, Daniel Levy, Daniel Gonz'alez, Danny Hernandez, Danqi Chen,
  Daphne Ippolito, Dar Gilboa, David Dohan, D.~Drakard, David Jurgens,
  Debajyoti Datta, Deep Ganguli, Denis Emelin, Denis Kleyko, Deniz Yuret, Derek
  Chen, Derek Tam, Dieuwke Hupkes, Diganta Misra, Dilyar Buzan, Dimitri~Coelho
  Mollo, Diyi Yang, Dong-Ho Lee, Ekaterina Shutova, Ekin~Dogus Cubuk, Elad
  Segal, Eleanor Hagerman, Elizabeth Barnes, Elizabeth~P. Donoway, Ellie
  Pavlick, Emanuele Rodol{\`a}, Emma~FC Lam, Eric Chu, Eric Tang, Erkut Erdem,
  Ernie Chang, Ethan~A. Chi, Ethan Dyer, Ethan Jerzak, Ethan Kim, Eunice~Engefu
  Manyasi, Evgenii Zheltonozhskii, Fan Xia, Fatemeh Siar, Fernando
  Mart'inez-Plumed, Francesca Happ'e, François Chollet, Frieda Rong, Gaurav
  Mishra, Genta~Indra Winata, Gerard de~Melo, Germ{\'a}n Kruszewski,
  Giambattista Parascandolo, Giorgio Mariani, Gloria Wang, Gonzalo
  Jaimovitch-L'opez, Gregor Betz, Guy Gur-Ari, Hana Galijasevic, Han~Sol Kim,
  Hannah Rashkin, Hanna Hajishirzi, Harsh Mehta, Hayden Bogar, Henry Shevlin,
  Hinrich Sch{\"u}tze, Hiromu Yakura, Hongming Zhang, Hubert Wong, Ian Aik-Soon
  Ng, Isaac Noble, Jaap Jumelet, Jack Geissinger, John Kernion, Jacob Hilton,
  Jaehoon Lee, Jaime~Fern{\'a}ndez Fisac, J.~Brooker Simon, James Koppel, James
  Zheng, James Zou, Jan Koco'n, Jana Thompson, Jared Kaplan, Jarema Radom,
  Jascha~Narain Sohl-Dickstein, Jason Phang, Jason Wei, Jason Yosinski,
  Jekaterina Novikova, Jelle Bosscher, Jenni Marsh, Jeremy Kim, Jeroen Taal,
  Jesse Engel, Jesujoba~Oluwadara Alabi, Jiacheng Xu, Jiaming Song, Jillian
  Tang, Jane~W Waweru, John Burden, John Miller, John~U. Balis, Jonathan
  Berant, Jorg Frohberg, Jos Rozen, Jos{\'e} Hern{\'a}ndez-Orallo, Joseph
  Boudeman, Joseph Jones, Joshua~B. Tenenbaum, Joshua~S. Rule, Joyce Chua,
  Kamil Kanclerz, Karen Livescu, Karl Krauth, Karthik Gopalakrishnan, Katerina
  Ignatyeva, Katja Markert, Kaustubh~D. Dhole, Kevin Gimpel, Kevin~Ochieng’
  Omondi, Kory~Wallace Mathewson, Kristen Chiafullo, Ksenia Shkaruta, Kumar
  Shridhar, Kyle McDonell, Kyle Richardson, Laria Reynolds, Leo Gao, Li~Zhang,
  Liam Dugan, Lianhui Qin, Lidia Contreras-Ochando, Louis-Philippe Morency,
  Luca Moschella, Luca Lam, Lucy Noble, Ludwig Schmidt, Luheng He,
  Luis~Oliveros Col'on, Luke Metz, Lutfi~Kerem cSenel, Maarten Bosma, Maarten
  Sap, Maartje ter Hoeve, Madotto Andrea, Maheen~Saleem Farooqi, Manaal
  Faruqui, Mantas Mazeika, Marco Baturan, Marco Marelli, Marco Maru,
  M~Quintana, Marie Tolkiehn, Mario Giulianelli, Martha Lewis, Martin Potthast,
  Matthew Leavitt, Matthias Hagen, M'aty'as Schubert, Medina Baitemirova,
  Melissa Arnaud, Melvin~Andrew McElrath, Michael~A. Yee, Michael Cohen, Mi~Gu,
  Michael~I. Ivanitskiy, Michael Starritt, Michael Strube, Michal Swkedrowski,
  Michele Bevilacqua, Michihiro Yasunaga, Mihir Kale, Mike Cain, Mimee Xu,
  Mirac Suzgun, Monica Tiwari, Mohit Bansal, Moin Aminnaseri, Mor Geva, Mozhdeh
  Gheini, T~MukundVarma, Nanyun Peng, Nathan Chi, Nayeon Lee, Neta Gur-Ari
  Krakover, Nicholas Cameron, Nicholas~S. Roberts, Nicholas Doiron, Nikita
  Nangia, Niklas Deckers, Niklas Muennighoff, Nitish~Shirish Keskar, Niveditha
  Iyer, Noah Constant, Noah Fiedel, Nuan Wen, Oliver Zhang, Omar Agha, Omar
  Elbaghdadi, Omer Levy, Owain Evans, Pablo Antonio~Moreno Casares, Parth
  Doshi, Pascale Fung, Paul~Pu Liang, Paul Vicol, Pegah Alipoormolabashi,
  Peiyuan Liao, Percy Liang, Peter~W. Chang, Peter Eckersley, Phu~Mon Htut,
  Pi-Bei Hwang, P.~Milkowski, Piyush~S. Patil, Pouya Pezeshkpour, Priti Oli,
  Qiaozhu Mei, QING LYU, Qinlang Chen, Rabin Banjade, Rachel~Etta Rudolph,
  Raefer Gabriel, Rahel Habacker, Ram'on~Risco Delgado, Rapha{\"e}l
  Milli{\`e}re, Rhythm Garg, Richard Barnes, Rif~A. Saurous, Riku Arakawa,
  Robbe Raymaekers, Robert Frank, Rohan Sikand, Roman Novak, Roman Sitelew,
  Ronan~Le Bras, Rosanne Liu, Rowan Jacobs, Rui Zhang, Ruslan Salakhutdinov,
  Ryan Chi, Ryan Lee, Ryan Stovall, Ryan Teehan, Rylan Yang, Sahib~J. Singh,
  Saif~M. Mohammad, Sajant Anand, Sam Dillavou, Sam Shleifer, Sam Wiseman,
  Samuel Gruetter, Sam Bowman, Samuel~S. Schoenholz, Sanghyun Han, Sanjeev
  Kwatra, Sarah~A. Rous, Sarik Ghazarian, Sayan Ghosh, Sean Casey, Sebastian
  Bischoff, Sebastian Gehrmann, Sebastian Schuster, Sepideh Sadeghi, Shadi~S.
  Hamdan, Sharon Zhou, Shashank Srivastava, Sherry Shi, Shikhar Singh, Shima
  Asaadi, Shixiang~Shane Gu, Shubh Pachchigar, Shubham Toshniwal, Shyam
  Upadhyay, Shyamolima Debnath, Siamak Shakeri, Simon Thormeyer, Simone Melzi,
  Siva Reddy, Sneha~Priscilla Makini, Soo hwan Lee, Spencer~Bradley Torene,
  Sriharsha Hatwar, Stanislas Dehaene, Stefan Divic, Stefano Ermon, Stella~Rose
  Biderman, Stephanie~C. Lin, Stephen Prasad, Steven~T. Piantadosi, Stuart~M.
  Shieber, Summer Misherghi, Svetlana Kiritchenko, Swaroop Mishra, Tal Linzen,
  Tal Schuster, Tao Li, Tao Yu, Tariq~A. Ali, Tatsuo Hashimoto, Te-Lin Wu, Theo
  Desbordes, Theodore Rothschild, Thomas Phan, Tianle Wang, Tiberius Nkinyili,
  Timo Schick, T.~N. Kornev, Timothy Telleen-Lawton, Titus Tunduny, Tobias
  Gerstenberg, Trenton Chang, Trishala Neeraj, Tushar Khot, Tyler~O. Shultz,
  Uri Shaham, Vedant Misra, Vera Demberg, Victoria Nyamai, Vikas Raunak,
  Vinay~Venkatesh Ramasesh, Vinay~Uday Prabhu, Vishakh Padmakumar, Vivek
  Srikumar, William Fedus, William Saunders, William Zhang, W~Vossen, Xiang
  Ren, Xiaoyu~F Tong, Xinyi Wu, Xudong Shen, Yadollah Yaghoobzadeh, Yair
  Lakretz, Yang Song, Yasaman Bahri, Ye~Ji Choi, Yichi Yang, Yiding Hao, Yifu
  Chen, Yonatan Belinkov, Yu~Hou, Yu~Hou, Yushi Bai, Zachary Seid, Zhao Xinran,
  Zhuoye Zhao, Zi~Fu Wang, Zijie~J. Wang, Zirui Wang, Ziyi Wu, Sahib Singh, and
  Uri Shaham. 2022.
\newblock \href {https://arxiv.org/abs/2206.04615} {Beyond the imitation game:
  Quantifying and extrapolating the capabilities of language models}.
\newblock \emph{ArXiv preprint}, abs/2206.04615.

\bibitem[{Vu et~al.(2020)Vu, Wang, Munkhdalai, Sordoni, Trischler,
  Mattarella-Micke, Maji, and Iyyer}]{Vu2020ExploringAP}
Tu~Vu, Tong Wang, Tsendsuren Munkhdalai, Alessandro Sordoni, Adam Trischler,
  Andrew Mattarella-Micke, Subhransu Maji, and Mohit Iyyer. 2020.
\newblock \href {https://doi.org/10.18653/v1/2020.emnlp-main.635} {Exploring
  and predicting transferability across {NLP} tasks}.
\newblock In \emph{Proceedings of the 2020 Conference on Empirical Methods in
  Natural Language Processing (EMNLP)}, pages 7882--7926, Online. Association
  for Computational Linguistics.

\bibitem[{Wang et~al.(2022)Wang, Xu, Fang, Liu, Sun, Xu, Zhu, and
  Zeng}]{Wang2022TrainingDI}
Shuohang Wang, Yichong Xu, Yuwei Fang, Yang Liu, Siqi Sun, Ruochen Xu,
  Chenguang Zhu, and Michael Zeng. 2022.
\newblock \href {https://doi.org/10.18653/v1/2022.acl-long.226} {Training data
  is more valuable than you think: A simple and effective method by retrieving
  from training data}.
\newblock In \emph{Proceedings of the 60th Annual Meeting of the Association
  for Computational Linguistics (Volume 1: Long Papers)}, pages 3170--3179,
  Dublin, Ireland. Association for Computational Linguistics.

\bibitem[{Wei et~al.(2021)Wei, Bosma, Zhao, Guu, Yu, Lester, Du, Dai, and
  Le}]{Wei2021FinetunedLM}
Jason Wei, Maarten Bosma, Vincent Zhao, Kelvin Guu, Adams~Wei Yu, Brian Lester,
  Nan Du, Andrew~M. Dai, and Quoc~V. Le. 2021.
\newblock \href {https://arxiv.org/abs/2109.01652} {Finetuned language models
  are zero-shot learners}.
\newblock \emph{ArXiv preprint}, abs/2109.01652.

\bibitem[{Ye et~al.(2021)Ye, Lin, and Ren}]{ye-etal-2021-crossfit}
Qinyuan Ye, Bill~Yuchen Lin, and Xiang Ren. 2021.
\newblock \href {https://doi.org/10.18653/v1/2021.emnlp-main.572}
  {{C}ross{F}it: A few-shot learning challenge for cross-task generalization in
  {NLP}}.
\newblock In \emph{Proceedings of the 2021 Conference on Empirical Methods in
  Natural Language Processing}, pages 7163--7189, Online and Punta Cana,
  Dominican Republic. Association for Computational Linguistics.

\end{thebibliography}
\bibliographystyle{acl_natbib}

\section*{Checklist}


\begin{enumerate}

\item For all authors...
\begin{enumerate}
  \item Do the main claims made in the abstract and introduction accurately reflect the paper's contributions and scope?
    \answerYes{}
  \item Did you describe the limitations of your work?
    \answerYes{Please see the experiment analysis sections and the appendix.}
  \item Did you discuss any potential negative societal impacts of your work?
    \answerYes{Please check the appendix.}
  \item Have you read the ethics review guidelines and ensured that your paper conforms to them?
    \answerYes{}
\end{enumerate}

\item If you are including theoretical results...
\begin{enumerate}
  \item Did you state the full set of assumptions of all theoretical results?
    \answerNA{}
	\item Did you include complete proofs of all theoretical results?
    \answerNA{}
\end{enumerate}

\item If you ran experiments...
\begin{enumerate}
  \item Did you include the code, data, and instructions needed to reproduce the main experimental results (either in the supplemental material or as a URL)?
    \answerYes{We include our code, data, and instructions in our supplemental material for reproducing all results presented in the paper. As the data is quite large, we also include a script to download it.}
  \item Did you specify all the training details (e.g., data splits, hyperparameters, how they were chosen)?
    \answerYes{We have specified the details of the base model (i.e., BART0) and the configurations for our retrieval-augmentation methods. Please check the experiment section and the appendix respectively.}
	\item Did you report error bars (e.g., with respect to the random seed after running experiments multiple times)?
    \answerYes{All our results are based on five different rounds, where each we use a different set of query examples. We also report multiple dimensions of the statistics in our table (e.g., the median, min, max, std, and mean).}
	\item Did you include the total amount of compute and the type of resources used (e.g., type of GPUs, internal cluster, or cloud provider)?
    \answerYes{We show these details in our appendix.}
\end{enumerate}

\item If you are using existing assets (e.g., code, data, models) or curating/releasing new assets...
\begin{enumerate}
  \item If your work uses existing assets, did you cite the creators?
    \answerYes{}
  \item Did you mention the license of the assets?
    \answerYes{}
  \item Did you include any new assets either in the supplemental material or as a URL?
    \answerYes{}
  \item Did you discuss whether and how consent was obtained from people whose data you're using/curating?
    \answerYes{The data we used are all open-source and publicly available via the ``datasets'' library from HuggingFace. }
  \item Did you discuss whether the data you are using/curating contains personally identifiable information or offensive content?
    \answerYes{Please refer to our appendix and the links to the used datasets.}
\end{enumerate}

\item If you used crowdsourcing or conducted research with human subjects...
\begin{enumerate}
  \item Did you include the full text of instructions given to participants and screenshots, if applicable?
    \answerNA{}
  \item Did you describe any potential participant risks, with links to Institutional Review Board (IRB) approvals, if applicable?
    \answerNA{}
  \item Did you include the estimated hourly wage paid to participants and the total amount spent on participant compensation?
    \answerNA{}
\end{enumerate}

\end{enumerate}

\clearpage


\appendix

\section*{Appendix (i.e., Supplementary Material) of ``Unsupervised Cross-Task Generalization via Retrieval Augmentation'' (Submission \# 2811)}

This appendix include more implementation details, additional experimental results for ablation studies, and more analysis as well as findings.
Please note that we have uploaded our code too (named ``ReCross'' folder).
We first further analyze the performance of ReCross with more ablation analysis in Sec.~\ref{app:analysis}, and present detailed case studies for specific datasets in Sec.~\ref{app:case}, and introduce more implementation details in Sec.~\ref{app:imp}.

\section{Additional analysis}
\label{app:analysis}

\subsection{Utility analysis by grouping upstream tasks.}
Table~\ref{tab:remove} shows the results of ReCorss under the scenarios where one specific group of upstream tasks are excluded from the index. 
This allows us to evaluate the impact of various upstream task categories on each downstream task.

\begin{table*}[h!]
\hspace{-0.7em}
\centering
\scalebox{0.75}{
\begin{tabular}{@{}r|c|cccccccc@{}}
\toprule
Task       & None              & $-$MCQA & $-$SUM.     & $-$EQA   & $-$Stmt.         & $-$CBQA  & $-$S2txt & $-$TopCls & $-$ParaIden \\ \midrule
ARC-c.   & 38.44$_{\pm0.99}$ & 39.36$_{\pm0.86}$    & 37.94$_{\pm1.51}$ & 39.54$_{\pm1.24}$ & 37.94$_{\pm1.51}$ & 39.32$_{\pm0.54}$ & 37.32$_{\pm1.77}$   & 37.94$_{\pm1.51}$     & 37.94$_{\pm1.51}$          \\
anli\_r3   & 35.76$_{\pm0.90}$ & 36.18$_{\pm0.88}$    & 36.90$_{\pm0.83}$ & 36.78$_{\pm1.04}$ & 35.72$_{\pm1.92}$ & 35.84$_{\pm2.35}$ & 37.42$_{\pm0.97}$   & 35.92$_{\pm1.32}$     & 36.42$_{\pm1.20}$          \\
hswag  & 47.28$_{\pm2.95}$ & 40.56$_{\pm8.71}$    & 49.28$_{\pm5.79}$ & 39.02$_{\pm7.49}$ & 46.46$_{\pm3.39}$ & 37.62$_{\pm5.98}$ & 46.00$_{\pm6.32}$   & 39.14$_{\pm7.50}$     & 44.34$_{\pm6.19}$          \\
obqa & 39.58$_{\pm2.80}$ & 36.12$_{\pm0.88}$    & 38.32$_{\pm2.33}$ & 38.52$_{\pm2.08}$ & 38.32$_{\pm2.33}$ & 35.98$_{\pm2.37}$ & 36.32$_{\pm2.86}$   & 38.32$_{\pm2.33}$     & 35.94$_{\pm1.70}$          \\
piqa       & 41.42$_{\pm1.02}$ & 39.60$_{\pm1.35}$    & 40.46$_{\pm2.08}$ & 41.64$_{\pm2.65}$ & 41.30$_{\pm2.47}$ & 41.56$_{\pm1.46}$ & 40.26$_{\pm2.17}$   & 40.42$_{\pm0.99}$     & 40.56$_{\pm0.80}$          \\
squad2  & 30.58$_{\pm1.61}$ & 31.70$_{\pm2.02}$    & 31.64$_{\pm1.63}$ & 33.10$_{\pm2.48}$ & 30.70$_{\pm1.61}$ & 31.06$_{\pm1.91}$ & 30.70$_{\pm1.61}$   & 31.60$_{\pm1.90}$     & 30.70$_{\pm1.61}$          \\
cb         & 44.79$_{\pm3.36}$ & 49.36$_{\pm3.55}$    & 44.50$_{\pm4.52}$ & 43.93$_{\pm3.26}$ & 40.79$_{\pm3.05}$ & 44.00$_{\pm5.42}$ & 43.36$_{\pm4.15}$   & 42.36$_{\pm7.36}$     & 40.50$_{\pm5.62}$          \\
wic        & 50.58$_{\pm0.24}$ & 49.82$_{\pm1.12}$    & 49.96$_{\pm0.93}$ & 50.08$_{\pm0.96}$ & 48.96$_{\pm2.47}$ & 48.90$_{\pm2.16}$ & 50.30$_{\pm0.79}$   & 49.74$_{\pm0.73}$     & 49.42$_{\pm0.92}$          \\
wsc        & 61.46$_{\pm1.47}$ & 58.04$_{\pm2.78}$    & 60.23$_{\pm2.66}$ & 60.54$_{\pm1.23}$ & 58.85$_{\pm3.67}$ & 59.19$_{\pm2.47}$ & 59.69$_{\pm2.21}$   & 60.19$_{\pm1.45}$     & 59.54$_{\pm3.27}$          \\
wngrnd & 55.46$_{\pm0.88}$ & 53.30$_{\pm1.52}$    & 52.34$_{\pm3.94}$ & 51.00$_{\pm4.94}$ & 54.44$_{\pm3.12}$ & 53.82$_{\pm2.59}$ & 52.20$_{\pm5.32}$   & 52.20$_{\pm3.33}$     & 50.74$_{\pm3.96}$     \\ \midrule \midrule
@mean & 44.53$_{\pm0.42}$ &	43.40$_{\pm0.92}$ &	44.16$_{\pm0.47}$ &	43.41$_{\pm1.20}$ &	43.35$_{\pm0.89}$ &	42.73$_{\pm0.75}$ &	43.36$_{\pm1.08}$ &	42.78$_{\pm1.38}$ &	42.61$_{\pm0.96}$ \\
\bottomrule

\end{tabular}
}
\caption{Performance on each downstream task when a given category of upstream tasks is removed from the upstream dataset and prevented from being retrieved. The column names are the task group names: MCQA=Multiple-Choice QA, SUM=Summarization, EQA=Extractive QA, Stmt.=Sentiment analysis, CBQA=closed-book QA, S2txt=structure-to-text, TopCls=Topic Classification, and ParaIden=Paraphrase Identification.}
\label{tab:remove}
\end{table*}

Our key findings are as follows: 
\begin{itemize}
    \item (1) Using all upstream tasks leads to the best overall performance, although for many target tasks there are some particular groups that are less useful than others. The last row shows this result and the summarization is the least useful group of upstream tasks.
    \item (2) The potential best performance of retrieval-augmentation methods can be even higher. That is, if we have an enhanced version of ReCross that can avoid examples from less useful groups, then the final performance can be even higher. For example, if ReCross were able to ignore MCQA examples for ARC task during retrieval augmentation, then the overall performance of ReCross can be even higher.
    \item (3) The utility analysis via grouping upstream tasks by their original task formulations does not align with general intuition. For example, people may think that MCQA (multiple-choice QA) should be more useful than other groups for the task of ARC, which is also a multiple-choice QA dataset. However, removing MCQA doesn't hurt the performance of ARC. Instead, it actually improves the performance by 1 point. We argue that the example-based utility is of more importance for analysis. 
\end{itemize}

\subsection{Template Perturbation}

\begin{figure*}[t]
	\hspace{-2em}
	\includegraphics[width=1.0\linewidth]{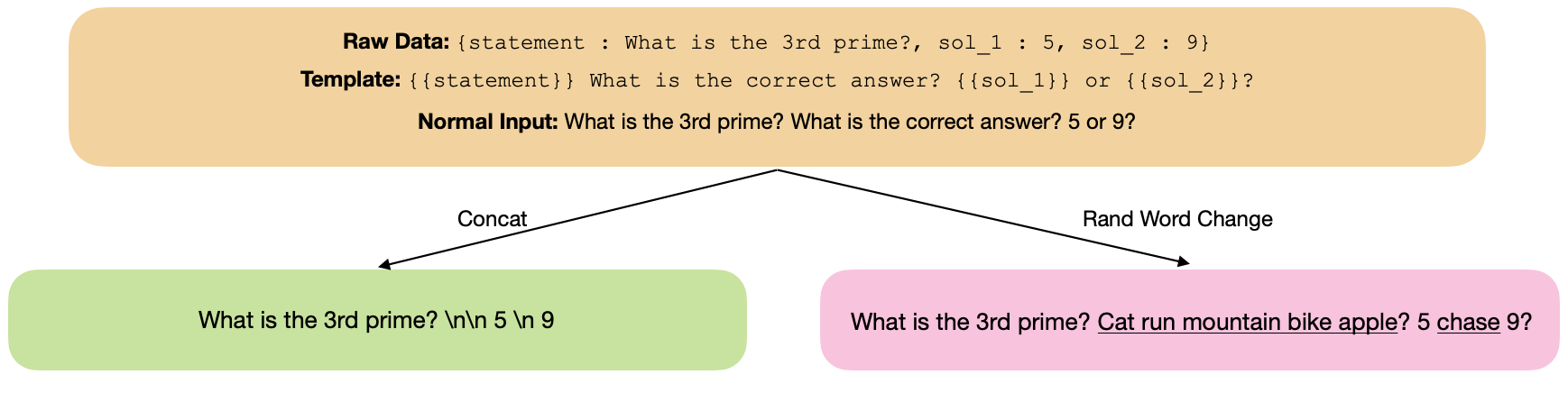} 
	\caption{Example of concatenation and random word change perturbation.}
	\label{fig:perturb_demo}
\end{figure*}

To investigate the importance of templates in retrieval quality, we investigated two methods of perturbing the templates of query examples $Q$: 1. Simply concatenate the elements in the raw data. 2. Change the words in the templates to random words to remove the semantic meaning. See figure~\ref{fig:perturb_demo} for an example. We than used these updated query examples and the same setup and configurations described in Section~\ref{ssec:setup_config} to perform unsupervised cross-task generalization. 


Table~\ref{tab:temp_perturb} shows that when we simply concatenate the elements in raw data, the performance degrades to a level close to random retrieval. On the other hand, if we construct the query examples as specified by the templates, even if we break the semantics of the template, the performance boost is largely preserved. This might mean that the formatting of input, for example the existence of parallel choices in some form, potentially plays an important role in the performance gain.

\begin{table*}[t]
\vspace{-1em}
\hspace{-0.8em}
\centering
\scalebox{0.8}{
\begin{tabular}{r|c|c||c|c|c|c||c|c}
\toprule
Target Task     & T0-3B  & {\underline{BART0}}   & Random            & SBERT             & ReCross$^\dag$     &{\underline{ReCross}} & \textbf{Concat} & \textbf{Change}  \\
\midrule
anli\_r3         & 26.00 & 30.50 & 35.34$_{\pm1.52}$ & 32.64$_{\pm2.53}$ & 36.70$_{\pm0.53}$ & 35.76$_{\pm0.90}$ & 34.14$_{\pm2.24}$ & 32.84$_{\pm6.33}$           \\
h-swag        & 34.40 & 39.40 & 33.84$_{\pm5.59}$ & 30.92$_{\pm7.82}$ & 44.36$_{\pm3.07}$ & 47.28$_{\pm2.95}$ & 35.74$_{\pm5.06}$ & 35.40$_{\pm10.82}$         \\
cb               & 53.93 & 39.64 & 47.07$_{\pm1.25}$ & 48.00$_{\pm3.28}$ & 44.50$_{\pm4.20}$ & 44.79$_{\pm3.36}$ & 39.29$_{\pm3.48}$ & 44.00$_{\pm{5.36}}$              \\
wic              & 45.70 & 46.70 & 41.04$_{\pm2.18}$ & 46.78$_{\pm2.22}$ & 49.90$_{\pm0.50}$ & 50.58$_{\pm0.24}$ & 46.88$_{\pm2.93}$ & 47.32$_{\pm1.91}$              \\
wsc              & 50.00 & 57.88 & 52.50$_{\pm2.29}$ & 52.69$_{\pm6.13}$ & 59.27$_{\pm1.96}$ & 61.46$_{\pm1.47}$ & 52.31$_{\pm5.17}$ & 57.31$_{\pm1.75}$   \\
winogrande       & 47.60 & 51.10 & 52.68$_{\pm0.83}$ & 52.18$_{\pm3.20}$ & 54.60$_{\pm1.35}$ & 55.46$_{\pm0.88}$ &52.28$_{\pm0.57}$ & 54.76$_{\pm2.07}$              \\ 
arc-chan.         & 41.30 & 35.70 & 33.28$_{\pm1.50}$ & 37.90$_{\pm1.22}$ & 37.78$_{\pm0.73}$ & 38.44$_{\pm0.99}$ &37.92$_{\pm0.48}$ & 38.24$_{\pm1.20}$               \\
obqa       & 38.50 & 34.40 & 28.72$_{\pm2.46}$ & 33.28$_{\pm1.24}$ & 36.98$_{\pm1.55}$ & 39.58$_{\pm2.80}$ &36.12$_{\pm3.14}$ &38.56$_{\pm2.06}$               \\
piqa             & 45.30 & 36.10 & 37.00$_{\pm2.71}$ & 38.54$_{\pm2.17}$ & 41.34$_{\pm1.75}$ & 41.42$_{\pm1.02}$ &39.76$_{\pm0.99}$ & 42.16$_{\pm1.86}$               \\
squadv2        & 30.60 & 32.40 & 29.86$_{\pm5.46}$ & 29.46$_{\pm0.84}$ & 30.26$_{\pm1.54}$ & 30.58$_{\pm1.61}$ & 30.74$_{\pm1.66}$  & 30.10$_{\pm1.22}$           \\
\midrule \midrule
All@mean   & 41.33 & 40.38 & 39.13$_{\pm2.06}$ & 40.24$_{\pm1.61}$ & 43.57$_{\pm0.68}$ & 44.53$_{\pm0.42}$ &40.52$_{\pm1.2}$ & 42.07$_{\pm1.5}$            \\
@median & 41.33 & 40.38 & 39.93           & 40.91           & 43.43           & 44.31     & 40.96  &41.69               \\
@min    & 41.33 & 40.38 & 35.66           & 38.28           & 42.65           & 44.16     & 38.77 & 40.37                 \\
@max    & 41.33 & 40.38 & 40.59           & 41.76           & 44.51           & 45.07    & 41.61 &44.33  \\             
\bottomrule

\end{tabular}
}
\caption{Two methods of template perturbation (concatenation and random word change) compared with main experiment results.}
\label{tab:temp_perturb}
\end{table*}
\begin{wraptable}{r}{7cm}
\vspace{-1cm}
\begin{minipage}{\textwidth}
\resizebox{0.5\textwidth}{!}{
\begin{tabular}{r|cc|cc}
\toprule
  Target Task    & Random    &\textbf{\underline{Random}+RR}  & SBERT   &\textbf{\underline{SBERT}+RR}    \\
  \midrule
  anli\_r3      &  35.34$_{\pm1.52}$ &31.58$_{\pm4.39}$ & 32.64$_{\pm2.53}$ & 28.10$_{\pm4.83}$\\
  h-swag     &    33.84$_{\pm5.59}$ & 33.20$_{\pm9.86}$ & 30.92$_{\pm7.82}$   & 37.80$_{\pm6.92}$\\
  cb           &    47.07$_{\pm1.25}$  &40.71$_{\pm1.84}$ & 48.00$_{\pm3.28}$  & 40.86$_{\pm7.80}$\\
  wic         &      41.04$_{\pm2.18}$  & 44.74$_{\pm0.88}$&46.78$_{\pm2.22}$  & 45.88$_{\pm2.19}$\\
  wsc        &       52.50$_{\pm2.29}$ & 50.38$_{\pm6.03}$& 52.69$_{\pm6.13}$  & 55.42$_{\pm2.66}$\\
  winogrande    &    52.68$_{\pm0.83}$  & 49.44$_{\pm13.80}$& 52.18$_{\pm3.20}$  & 53.02$_{\pm3.49}$\\ 
  arc-chan.    &      33.28$_{\pm1.50}$  & 33.52$_{\pm3.76}$& 37.90$_{\pm1.22}$  & 37.54$_{\pm1.87}$\\
  obqa  &     28.72$_{\pm2.46}$  & 25.96$_{\pm6.53}$& 33.28$_{\pm1.24}$  & 35.08$_{\pm3.27}$\\
  piqa      &       37.00$_{\pm2.71}$  &35.22$_{\pm5.25}$ & 38.54$_{\pm2.17}$   & 38.82$_{\pm2.06}$\\
  squadv2     &    29.86$_{\pm5.46}$  & 25.28$_{\pm3.93}$& 29.46$_{\pm0.84}$   & 29.56$_{\pm1.40}$\\
  \midrule \midrule
  All@mean &    39.13$_{\pm2.06}$ & 37.00$_{\pm2.91}$& 40.24$_{\pm1.61}$ & 40.21$_{\pm1.83}$\\
  @median &   39.93      & 37.06      & 40.91    & 39.81    \\
  @min  &   35.66      & 33.32 &   38.28     & 38.45    \\
  @max  &   40.59     & 40.26 &    41.76      & 42.82   \\             
  \bottomrule
  
\end{tabular}
}
\end{minipage}
  \caption{Random and SBERT with Re-Ranking (RR) (bold font columns)}
  \label{tab:random_sbert_rerank}
\vspace{-1cm}
\end{wraptable}

  

\subsection{Re-ranking for Random and SBERT}
We evaluated training re-rankers for random and SentenceBERT retrievers. Specifically, we applied the same distant supervision mining methods introduced in Section~\ref{ssec:ds} on data retrieved by \textbf{Random} and \textbf{SBERT}. Table~\ref{tab:random_sbert_rerank} shows the results.
We can see that reranking does not improve the results for both Random and SBERT retriever. We believe it is because that the initial retrieval results are not good enough, so that the distant supervision mined from them are thus also not of good quality. 

\subsection{Mining distant supervision for multiple iterations.}
The algorithm that we proposed in Alg.~\ref{alg:ds_reranker} can be extended to an iterative process. That is,
we can continually update the reranker module and uses the retrieved results from the latest reranker to mine the training data for the next iteration.
Although this self-training style process sounds promising, our empirical results show that the overall performance starts to saturate after the first iteration and using the 2nd-iteration re-ranker won't improve the overall performance anymore. 
We think there can be better methods of continual learning to obtain a reranker module for better performance, while it is beyond the scope of this work. 
We hope this can be a promising future direction.

\subsection{Transferring ReCross for Larger Base Models.}
Recall that we choose to use BART0 as our base model for its smaller size and comparable results.
People may wonder what if we transfer the ReCross methods for larger base models. 
Therefore, we conduct a pilot study on this. 
Considering the size of T0, we choose to only fine-tune its last few layers of T0-3B and still use the prior materials from BART0 (i.e., the BART0-based index and the trained reranker). We found that the performance is not improved over simply using T0-3B for zero-shot inference.
We conjecture there are two major reasons for this: 1) the parameter-efficient tuning method need to added here to improve the training efficiency, 2) the BART0-based index and the associated reranker do not align with the other models such as T0-3B.
We admit this could be one limitation of our methods -- i.e., the index and reranker are specific to the base model that is used to generate them. 
In order to address these challenges, we argue that studying the common space of the index created by different encoders will be an important direction.


\section{Case studies}
\label{app:case}
In this section, we discuss two specific datasets with detailed analysis as they have quite special results in Table~\ref{tab:main} and Table~\ref{tab:remove}.

\subsection{SuperGLUE CommitmentBank (cb)}
For the SuperGLUE CommitmentBank dataset, instances retrieved by the BART retriever are predominantly multiple choice question-answering. However, heat map and remove-one-group analysis shows that re-training on instances from multiple choice question-answering seems to undermine the model’s zero-shot performance on this dataset. We examined the output of the model and discovered that the model tends to make one type of error a lot more often when re-trained using multiple choice question-answering: instead of answering yes, no, possible, or impossible, it picks part of the discourse as its prediction. 

For example:

Input: ``Suppose A: I’m like, I’ll get a job some day and my boss will pay for it, I’ll be needed. B: Yeah. A: Because, um, I didn’t want to go do it myself because I didn’t think I was really going to use it. Can we infer that ``he was really going to use it''? Yes, no, or maybe?''

Output: ``A: I didn’t want to go do it myself because I didn't think I was really going to use it.''

We believe this is because the model misunderstood the people having the discourse (A and B) to be the options for answers. The abundance of the template of ``A:xxx, B:xxx'' in the SuperGLUE CommitmentBank dataset might be the reason why the BART retriever retrieved mostly from multiple choice question-answering in the first place.

\subsection{SQuAD V2}
For the dataset SQuAD V2, the retriever typically finds upstream examples from extractive question answering datasets, which match the format of SQuAD V2 inputs closely. However, we find that when we exclude extractive question answering examples from the upstream dataset, performance on SQuAD V2 improves. To explain this unexpected result, we note that the majority of our test examples for SQuAD V2, despite being formatted as extractive question answering tasks, are examples which expect the model to output whether or not the question is answerable. The ‘context and question’ format of the SQuAD V2 examples causes the retriever to focus on extractive question answering examples, but because most of the examples focus on answerability (a distinct task from extractive question answering), these examples are not helpful. 

We speculate that by excluding extractive question answering from the upstream dataset, the model avoids these misleading irrelevant examples and is able to retrieve more related examples for determining if a question is answerable. For example, our results show that when extractive question answering examples are excluded, the retriever finds examples from tasks such as Wiki QA, which asks whether or not a proposed answer is a valid answer to a given question (a more relevant task to determining if a question is answerable).

\section{Implementation details}
\label{app:imp}

\subsection{Retrieval aggregation.}
\label{app:ret_agg}
Note that the target size of our retrieved data is $|R|$ and we have $|Q|$ query examples. To retrieve $|R|$ examples, we search for the top-$K$ examples for each query example, where $K=\ceil{\frac{|R|}{|Q|}}$, and then take the first $|R|$ of them when $K |Q| > |R|$. 
Our results have shown that this method is more effective than other strategies, such as combining the distance scores generated for each query example. 
Note that by retrieving the top-$K$ examples, we may repeat examples that are close to multiple query vectors. 
This effect is desirable because it allows us to naturally focus more on the especially relevant upstream examples in re-learning.


\subsection{Upstream learning.}
\paragraph{Upstream tasks.}
Here we refer to the T0's paper (cited in our main paper) for Figure~\ref{fig:upstream}, which shows the list of upstream tasks and their categories. We use this taxonomy to conduct ablation study.
Please find the link to download these datasets from huggingface/dataset from our submitted code. All datasets are publicly available and their license are suitable for open-source research. 
We do not see any ethical concerns from using such datasets for learning a model and developing the ReCross method to further improve their task generalization performance.

\paragraph{Training details.}
We specify the hyper-parameters and the concrete for training the BART0 models in our submitted code. Please read the ``Readme.md'' file where we point to the script and configurations for training BART0. 
Our GPU type is Quadro RTX 6000 and 8000. 

\subsection{Retrieval Methods.}
Similarly, we leave the details such as the hyper-parameters and the concrete pipeline for running the retrieval augmentation methods (i.e., ReCross and the other baseline methods) in a unified framework that is presented in our code.

\section{{Others}}

\subsection{Evaluation metrics.}
\label{app:EM}

\paragraph{Results with the standard EM.}
In Table~\ref{tab:main_EM}, we report our main experimental results (the equivalent results to those in Table~\ref{tab:main}) with the standard EM metric instead of the SoftEM metric used in Table~\ref{tab:main}. We can see that the relative performance from the ReCross framework is about the same as in Table~\ref{tab:main}, although the absolute numbers are mostly smaller due to a more strict matching by EM. 

\begin{table*}[h]
\hspace{-0.8em}
\centering
\scalebox{0.98}{
\begin{tabular}{r|c|c||c|c|c|c||c}
\toprule
Target Task     & T0-3B  & \textbf{\underline{BART0}}   & Random            & SBERT             & ReCross$^\dag$     &\textbf{\underline{ReCross}}   & $\Delta$ \\
\midrule
anli\_r3        & 24.30 & 24.30 & 27.80$_{\pm2.12}$ & 25.62$_{\pm2.35}$ & 31.02$_{\pm0.87}$ & 30.18$_{\pm1.48}$ & 5.88  \\
h-swag        & 22.20 & 24.20 & 26.04$_{\pm2.61}$ & 22.88$_{\pm2.44}$ & 27.48$_{\pm2.04}$ & 26.04$_{\pm1.19}$ & 1.84  \\
cb               & 49.29 & 26.79 & 31.64$_{\pm3.27}$ & 34.21$_{\pm5.12}$ & 30.00$_{\pm2.65}$ & 31.57$_{\pm6.18}$ & 4.79  \\
wic              & 44.70 & 45.80 & 45.26$_{\pm4.13}$ & 46.78$_{\pm2.22}$ & 49.90$_{\pm0.50}$ & 50.58$_{\pm0.24}$ & 4.78  \\
wsc             & 48.85 & 54.42 & 53.96$_{\pm3.29}$ & 52.42$_{\pm6.09}$ & 59.15$_{\pm1.82}$ & 61.42$_{\pm1.51}$ & 7.00  \\
winogrande       & 47.00 & 49.50 & 50.44$_{\pm0.57}$ & 50.80$_{\pm2.89}$ & 54.16$_{\pm1.18}$ & 54.42$_{\pm1.10}$ & 4.92  \\
arc-chan.        & 32.10 & 23.70 & 26.84$_{\pm1.37}$ & 27.02$_{\pm2.52}$ & 26.86$_{\pm1.90}$ & 27.16$_{\pm1.78}$ & 3.46  \\
obqa       & 38.80 & 34.10 & 27.20$_{\pm1.24}$ & 33.76$_{\pm1.51}$ & 36.90$_{\pm2.56}$ & 39.56$_{\pm2.79}$ & 5.46  \\
piqa           & 33.40 & 29.10 & 29.32$_{\pm3.26}$ & 28.94$_{\pm3.08}$ & 31.70$_{\pm3.17}$ & 30.46$_{\pm2.34}$ & 1.36  \\
squadv2        & 23.70 & 26.30 & 24.20$_{\pm4.34}$ & 21.90$_{\pm1.17}$ & 22.96$_{\pm1.95}$ & 23.32$_{\pm2.16}$ & -2.98 \\
\midrule \midrule
All@mean   & 36.43 & 33.82 & 34.27$_{\pm1.66}$           & 34.43$_{\pm1.14}$           & 37.01$_{\pm0.94}$           & 37.47$_{\pm0.73}$           & 3.65  \\
@median  & 36.43 & 33.82 & 34.90           & 34.91           & 36.62           & 37.17           & 2.34  \\
@min   & 36.43 & 33.82 & 31.33           & 32.91           & 36.22           & 36.93           & 1.05  \\
@max    & 36.43 & 33.82 & 35.35           & 35.79           & 38.41           & 38.75           & 1.70   \\             
\bottomrule

\end{tabular}
}
\caption{\textbf{The main experimental results (\%) for unsupervised cross-task generalization in the \textbf{standard EM} metric, i.e., the EM version of Table~\ref{tab:main}.}
}
\label{tab:main_EM}
\end{table*}


\subsection{Empirical results for few-shot learning.}

We show the empirical results related to the few-shot setting in Table~\ref{tab:fs}.

\begin{table*}[h]
\centering
\scalebox{0.98}{
\begin{tabular}{r|c|c||c|c|c}
\toprule
Target Task     & \textbf{\underline{BART0}}   & ReCross (ReX)    & Few-Shot(FS)             & FS+ReX(Mix) & FS+ReX(2-stage) \\
\midrule
anli\_r3   & 30.50 & 38.44$_{\pm0.99}$ & 34.59$_{\pm2.33}$ & 35.71$_{\pm1.59}$ & 36.26$_{\pm1.48}$   \\
h-swag     & 39.40 & 35.76$_{\pm0.90}$ & 42.61$_{\pm2.15}$ & 44.04$_{\pm3.60}$ & 43.99$_{\pm1.92}$   \\
cb         & 39.64 & 47.28$_{\pm2.95}$ & 52.57$_{\pm6.11}$ & 62.64$_{\pm5.68}$ & 65.36$_{\pm6.70}$   \\
wic        & 46.70 & 39.58$_{\pm2.80}$ & 48.22$_{\pm2.10}$ & 49.23$_{\pm1.52}$ & 48.21$_{\pm2.57}$   \\
wsc        & 57.88 & 41.42$_{\pm1.02}$ & 53.15$_{\pm3.80}$ & 55.65$_{\pm7.82}$ & 54.54$_{\pm5.22}$   \\
winogrande & 51.10 & 30.58$_{\pm1.61}$ & 54.24$_{\pm1.57}$ & 53.24$_{\pm1.81}$ & 53.87$_{\pm1.72}$   \\
arc-chan.  & 35.70 & 44.79$_{\pm3.36}$ & 36.36$_{\pm2.20}$ & 36.34$_{\pm2.64}$ & 37.50$_{\pm2.94}$   \\
obqa       & 34.40 & 50.58$_{\pm0.24}$ & 34.49$_{\pm4.21}$ & 38.45$_{\pm2.68}$ & 37.15$_{\pm2.63}$   \\
piqa       & 36.10 & 61.46$_{\pm1.47}$ & 47.38$_{\pm4.58}$ & 51.93$_{\pm2.72}$ & 52.08$_{\pm1.95}$   \\
squadv2    & 32.40 & 55.46$_{\pm0.88}$ & 41.92$_{\pm6.68}$ & 51.30$_{\pm3.23}$ & 50.38$_{\pm6.46}$   \\
\midrule \midrule
All@mean   & 40.38 & 44.54         & 44.55         & 47.85         & 47.93    \\
\bottomrule

\end{tabular}
}
\caption{\textbf{The few-shot related empirical results in SoftEM.}}
\label{tab:fs}
\end{table*}

\paragraph{Experimental setup.} We assume that the labels of the examples in the query set are available, and directly use them to fine-tune the upstream model for learning the target task. We tune the hyper-parameters (epochs and learning rates) such that they do not overfit the few-shot data and lead to a better performance over BART0. Note that the real performance of few-shot learning performance may be lower than the ones in the table because there is not enough development data for us to tune hyper-parameters for each target task. 

\paragraph{Few-shot learning is not even better than the unsupervised ReCross.} Although FS can outperform ReCross in some target tasks, the two approaches have very similar overall performance on 10 tasks. Even in such an unfair setting, ReCross shows great benefits to the users.

\paragraph{ReCross and Few-Shot together can produce better performance.} We attempted to use both the few-shot data and the retrieved data for generalization. The FS+RC(mix) method simply merge the 16 labeled query examples (i.e., few-shot) and the 512 retrieved data (by ReCross) to get a larger dataset for fine-tuning BART0. The  FS+RC(2-stage) method updates the model firstly with the 512 retrieved data and then train the fine-tuned model with the 16 FS data. Both methods show a great enhancement over FS and RC used separately. This is to say, RC is still beneficial in the FS setting.

\end{document}